\documentclass{article}
\usepackage{ijcai21}

\usepackage{times}
\usepackage{soul}
\usepackage{url}

\usepackage[hidelinks]{hyperref}
\usepackage[utf8]{inputenc}
\usepackage[small]{caption}
\usepackage{graphicx}
\usepackage{amsmath}
\usepackage{amsfonts}
\usepackage{amsthm}
\usepackage{booktabs}
\usepackage{subfigure}
\usepackage{float}
\usepackage{setspace}
\usepackage{tabularx}
\usepackage{textcomp}
\usepackage{verbatim}
\usepackage{array}
\usepackage{multirow}
\usepackage{lineno}
\usepackage[ruled,vlined]{algorithm2e}
\UseRawInputEncoding

\title{Graph Entropy Guided Node Embedding Dimension Selection \mbox{for Graph Neural Networks}}

\author{
Gongxu Luo$^{1,2}$\and
Jianxin Li$^{1,2}$\and
Jianlin Su$^3$\and
Hao Peng$^1$\and
Carl Yang$^4$\and
Lichao Sun$^5$\and \\
Philip S. Yu$^6$\And
Lifang He $^5$\\
\affiliations
$^1$Beijing Advanced Innovation Center for Big Data and Brain Computing, Beihang University, China\\
$^2$School of Computer Science and Engineering, Beihang University, China\\
$^3$ Shenzhen Zhuiyi Technology Co., Ltd, China \\
$^4$Department of Computer Science, Emory University, USA\\
$^5$Department of Computer Science and Engineering, Lehigh University, USA\\
$^6$Department of Computer Science, University of Illinois at Chicago, USA\\
\emails
\{luogx, lijx, penghao \}@act.buaa.edu.cn, bojonesu@wezhuiyi.com,
j.carlyang@emory.edu,
psyu@uic.edu,
\{lis221,lih319\}@lehigh.edu
}

\begin{document}

\maketitle
\newcolumntype{L}[1]{>{\raggedright\arraybackslash}p{#1}}
\newcolumntype{C}[1]{>{\centering\arraybackslash}p{#1}}
\newcolumntype{R}[1]{>{\raggedleft\arraybackslash}p{#1}}

\begin{abstract}
Graph representation learning has achieved great success in many areas, including e-commerce, chemistry, biology, etc.
However, the fundamental problem of choosing the appropriate dimension of node embedding for a given graph still remains unsolved.
The commonly used strategies for Node Embedding Dimension Selection~(NEDS) based on grid search or empirical knowledge suffer from heavy computation and poor model performance. In this paper, we revisit NEDS from the perspective of minimum entropy principle. Subsequently, we propose a novel Minimum Graph Entropy~(MinGE) algorithm for NEDS with graph data.
To be specific, MinGE considers both feature entropy and structure entropy on graphs, which are carefully designed according to the characteristics of the rich information in them.
The feature entropy, which assumes the embeddings of adjacent nodes to be more similar, connects node features and link topology on graphs. The structure entropy takes the normalized degree as basic unit to further measure the higher-order structure of graphs.
Based on them, we design MinGE to directly calculate the ideal node embedding dimension for any graph.
Finally, comprehensive experiments with popular Graph Neural Networks~(GNNs) on benchmark datasets demonstrate the effectiveness and generalizability of our proposed MinGE.
\end{abstract}

\section{Introduction}
In recent years, Graph Neural Networks~(GNNs)~\cite{wu2020comprehensive} have attracted tremendous attention from both research and industry, due to its powerful representation capability for large amounts of graph structured data in practice, \emph{e.g.,} social networks, citation networks, road networks.
GNNs are mostly used to compute distributed node representations~(\textit{a.k.a.}~embeddings), as dense vectors, which serves as the key to various downstream tasks in graph related applications~\cite{peng2019hierarchical,sun2021sugar}. 
The dimension of node embedding, as a crucial hyperparameter, has a significant influence on the performance of GNNs. First, node embeddings with too small dimensions limit the information expressiveness of nodes, whereas those with too large dimensions suffer from overfitting. Second, the number of parameters of GNNs that build on node embeddings is usually a linear or quadratic function of node embedding dimension, which directly affects model complexity and computational efficiency~\cite{DBLP:conf/nips/YinS18}.
\begin{figure}[!t]
 \centering
  \subfigtopskip=0pt
  \subfigbottomskip=-10pt
 \includegraphics[width=8.5cm, height=3.5cm]{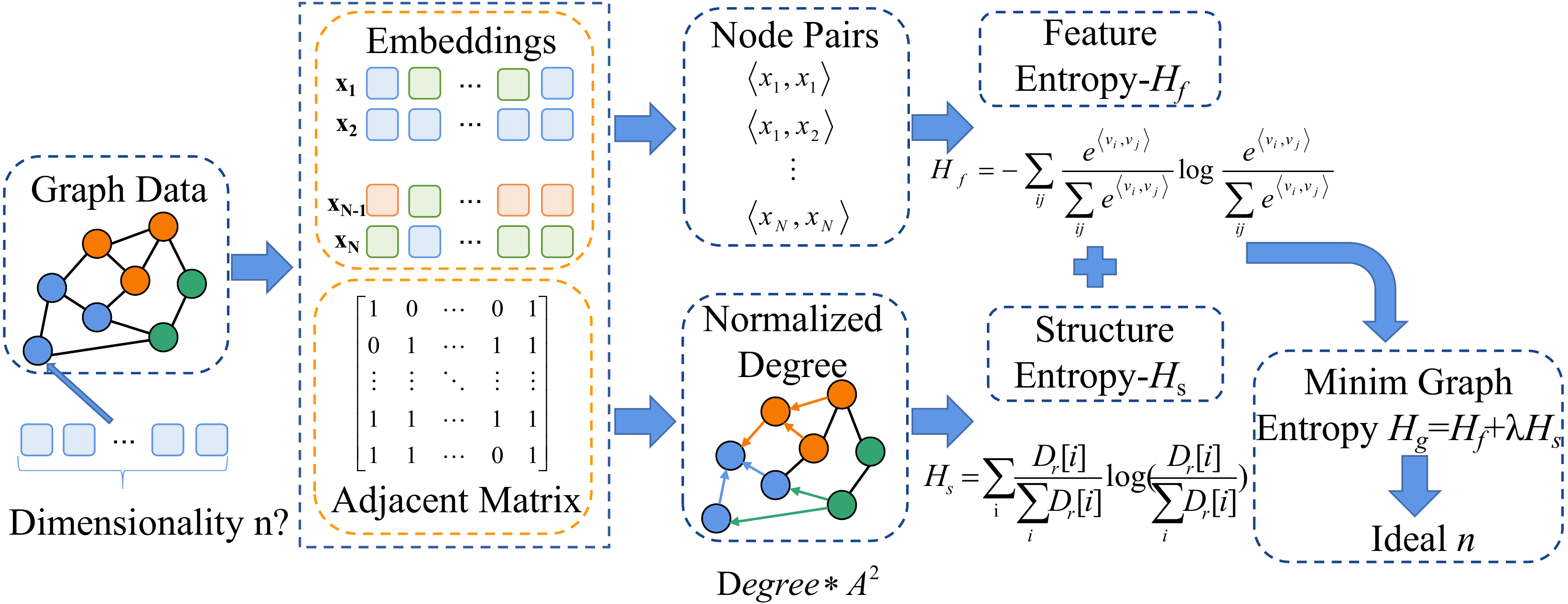}
 \caption{The overview of MinGE. The graph entropy considers both feature entropy and structure entropy to guide NEDS for a given graph. The feature entropy connects node features and link topology on graphs, while the structure entropy further measures the higher-order structure of graphs.}
 \label{fig1}
\end{figure}
However, we often have no idea towards the appropriate dimension of node embedding and have to rely on grid search or domain knowledge to adjust it as a model hyperparameter given a new graph. These approaches often suffer from the following problems: \textbf{1)} Choosing node embedding dimension by empirical experience can easily lead to poor model performances. \textbf{2)} Experimental strategies such as grid search are resource-consuming and time-consuming. This raises the questions of what is the appropriate dimension of node embedding for given graphs and what should we base on to choose such dimension?

Recently, several researchers~\cite{DBLP:journals/eaai/LiuSZOW20,DBLP:conf/ijcai/ZhuLYLGZ18,DBLP:journals/pr/ShenLBPZF20} have attempt to tackle these problems through dimensionality reduction. However, these methods focus on choosing important features or using complex matrix transformation to reduce from a large dimension instead of direct selection based on the graph. Moreover, different from sequence and image data, graph data contains rich link topology. The general dimensionality reduction methods only consider node features but ignore it for GNNs, which results in loss of useful information.

In this paper, inspired by recent dimension selection works~\cite{DBLP:conf/nips/YinS18,DBLP:journals/corr/abs-2008-07720,WinNT} in Natural Language Processing~(NLP), we revisit NEDS from the perspective of minimum entropy principle~\cite{zhu1997minimax}. 
Specifically, entropy is a natural way to measure the uncertainty of the dimension, the principle of minimum entropy motivates us to choose an ideal node embedding dimension for any given graphs by minimizing the uncertainty of the graph.
As shown in Fig.~\ref{fig1}, we design a novel graph entropy that contains feature entropy and structure entropy to account for the information regarding both node features and link structures on graphs.
Subsequently, we propose a novel Minimum Graph Entropy~(MinGE) algorithm\footnote{https://github.com/RingBDStack/MinGE} to directly select an appropriate node embedding dimension for a given graph through minimizing the graph entropy. As a consequence, MinGE leads to appropriate NEDS without the need of additional domain knowledge, and is obviously more efficient than grid search. 



The main contributions of this paper are summarized as follows:~\textbf{(1)} We stress the importance of direct NEDS for GNNs on arbitrary graphs and revisit it from an intuitive perspective based on the principle of minimum entropy.~\textbf{(2)} We develop the MinGE algorithm based on our novel design of graph entropy to enable direct NEDS, which considers the rich feature and structure information on graphs.~\textbf{(3)} Experimental results with several popular GNNs such as GCN, GAT, and GCNII over node classification and link prediction tasks on benchmark datasets demonstrate the effectiveness and generalizability of our proposed MinGE algorithm.

\section{Related Work}
\paragraph{Dimension Selection.}
How to choose the appropriate embedding dimension is always an open and challenging problem for deep learning. In NLP, motivated by the unitary-invariance of word embedding,~\cite{DBLP:conf/nips/YinS18} proposed the pairwise inner product loss to measure the dissimilarities between word embeddings. Inspired by the this work,~\cite{DBLP:conf/emnlp/Wang19a} proposed a fast and reliable method based on Principle Components Analysis~(PCA) to choose the dimension of word embedding, which use grid search to reduce the complexity of the algorithm. By rethinking the classical skip-gram algorithm, ~\cite{DBLP:journals/corr/abs-2008-07720} proposed a novel information criteria-based method to select the dimension of word embedding, and gave a theoretical analysis from the perspective of probability theory. Similarly, in Computer Vision~(CV), researchers also try to reduce the representation dimensionality by sampling operation on the input data~\cite{luo2020dynamically}.
In graph mining, existing work about embedding dimension selection is based on dimensionality reduction to select important features. Many previous dimensionality reduction algorithms are based on a predefined graph, however, there are noise and redundant information in original data.~\cite{DBLP:conf/ijcai/XiongNH17} proposed the linear manifold regularization with adaptive graph to directly incorporate the graph construction into the objective function to solve the problem. Similarly,~\cite{DBLP:journals/eaai/LiuSZOW20} proposed the termed discriminative sparse embedding to learn a sparse weight matrix to reduce the effects of redundant information and noise of original data. In order to avoid choosing the dimension that is outliers,~\cite{DBLP:conf/ijcai/ZhuLYLGZ18} proposed a robust graph dimensionality reduction algorithm to map high-dimension data into lower-dimensional intrinsic space by a transformation matrix. Although there are amounts of fantastic work for dimension selection. However, these methods focus on studying how to choose important features to reduce the dimensionality instead of direct selection based on the graph. Moreover, these methods ignore the rich link structures of graph data in dimensionality reduction, which results in loss of information.

\paragraph{Structure Entropy.}
The conception of structure entropy originate from the research of~\cite{DBLP:journals/tit/Shannon53b}, who establish the structural theory of information to support the analysis of communication  system. Subsequently, a large number of methods are proposed about understanding the complexity of networks.~\cite{DBLP:journals/entropy/MowshowitzD12} proposed the entropy of graph to measure the entropy of the distribution $p(G)$ at the global level. Besides, there were some methods to measure the structure entropy of nodes.~\cite{raychaudhury1984discrimination} first proposed the local measure of graph entropy, which is distance-based. Some extension works such as parametric graph entropy~\cite{DBLP:journals/amc/Dehmer08}, Gibbs entropy~\cite{bianconi2009entropy}, Shannon entropy, and Von Neumann entropy~\cite{braunstein2006laplacian} designed the structure information measurement of the network from different angles. Furthermore,~\cite{DBLP:journals/tit/LiP16} first proposed the metric for structure information and defined the K-dimensional structure information of the graph, which can not only detect the natural or true structure but also can measure the complexity of dynamic evolving networks.

So far, there is still no effective solution to directly select an appropriate node embedding dimension for any given graph. Therefore, in this paper, we revisit the NEDS from the perspective of minimum entropy principle. Based on it, we design a novel MinGE algorithm that contains feature entropy and structure entropy for direct NEDS.

\section{Methodology}
In this section, we first introduce the basic notations used in this paper. Then we present the overall framework of our minimum graph entropy approach MinGE, followed by the detail of how to calculate the feature entropy and the structure entropy for graph data with rich link structures.
\subsection{Notation}
Suppose we are given an undirected graph $G=(V,E,A)$, where $V$ denotes the node set of the graph, $E$ denotes the set of edges, and $A \in \mathbb{R}^{N \times N}$ denotes the adjacency matrix with the element $A[i, j] = A[j, i]$ indicating whether edge $(v_{i},v_{j})$ in $E$. $N$ is the number of nodes. The link structures can be presented by the first-order adjacency matrix $A$ and the second-order adjacency matrix $A^{2}$. The adjacency matrix after normalization is denoted as $A_{r}$. Similarly, the second-order adjacency matrix after normalization is denote as $A_{r}^{2}$. The degrees of all nodes are represented by a vector $D \in \mathbb{R}^{1 \times N}$, and the normalized degree is $D_{r}$.
\subsection{Graph Entropy}
Information entropy proposed by Shannon~\cite{DBLP:journals/bstj/Shannon48} is a measure of information uncertainty, which is defined as:
\begin{equation}
    H = -\sum_{i}^{n}P_{i}logP_{i},
\end{equation}
where $H$ denotes the entropy, $P_{i}$ is the probability of event $i$, $n$ is the number of events. It indicates that, the smaller the entropy, the lower information uncertainty, which equally means the more useful information. However, using single information entropy to measure graph data with rich link structures is brittle. Therefore, in this paper, we propose a novel graph entropy to consider both rich node features and link structures on graphs, which is defined as:
\begin{equation}
    H_{g} = H_{f} + \lambda H_{s},
\end{equation}
where $H_{g}$ represents graph entropy, $H_{f}$ and $H_{s}$ represents feature entropy and structure entropy respectively. $\lambda$ is a hyperparameter that controls the ratio of structure entropy for NEDS. In the following, we will discuss how to define $H_{f}$ and $H_{s}$ in detail. Especially for the feature entropy $H_{f}$, it establishes the connection between dimension $n$ and graph entropy. The structure entropy further search the ideal node embedding dimension. Finally, we can directly calculate the appropriate node embedding dimension $n$ by setting $H_{g}=0$.
\paragraph{Feature Entropy.}
Information entropy measures the uncertainty of information. The core of information entropy is how to define the basic unit of events. Inspired by the exploration of dimension selection in the field of NLP \cite{WinNT}, for graph data, we design the feature entropy as shown in Fig.~\ref{fig1}. Based on the assumption that the node embeddings of adjacent nodes to be more similar, we use node embedding dot product of node pairs as the basic unit and the probability is defined as:
\begin{equation}
    P(v_{i},v_{j})= \frac{e^{\langle v_{i},v_{j}\rangle}}{\sum_{i,j}e^{\langle v_{i},v_{j}\rangle}},
\end{equation}%
 where $v_{i}$ and $v_{j}$ are the corresponding node embedding, $\langle \cdot, \cdot \rangle$ is the dot product operation. For brevity, we denote $Z=\sum_{i,j}e^{\langle v_{i},v_{j}\rangle}$. The feature entropy of graph data is defined as:
 \begin{align}
    H_{f}&=-\sum_{ij}P(v_{i},v_{j})logP(v_{i},v_{j})\notag \\
    &=-\sum_{ij}\frac{e^{\langle v_{i},v_{j}\rangle}}{Z}log\frac{e^{\langle v_{i},v_{j}\rangle}}{Z}\notag\\
    &=logZ-\frac{1}{Z}\sum_{ij}e^{\langle v_{i},v_{j}\rangle}\langle v_{i},v_{j}\rangle.
 \end{align}
To calculate Eq.~(4), we approximate the sum operation with sampling\cite{WinNT}, which is defined as:
\begin{equation}
\begin{aligned}
    Z&=\sum_{ij}e^{\langle v_{i},v_{j}\rangle}=N^2\frac{1}{N^2}\sum_{ij}e^{\langle v_{i},v_{j}\rangle}\\
    &\approx N^2E_{v_{i},v_{j}}(e^{\langle v_{i},v_{j}\rangle}),
\end{aligned}
\end{equation}
\begin{equation}
    \begin{aligned}
    \sum_{ij}e^{\langle v_{i},v_{j}\rangle}\langle v_{i},v_{j}\rangle&=N^2\frac{1}{N^2}\sum_{ij}e^{\langle v_{i},v_{j}\rangle}\langle v_{i},v_{j}\rangle\\
    &\approx N^2E_{v_{i},v_{j}}(e^{\langle v_{i},v_{j}\rangle}\langle v_{i},v_{j}\rangle),
    \end{aligned}
\end{equation}
where $E$ is the expectation operation. Plugging the approximate expressions from Eqs.~(5) and (6) into Eq.~(4), we can calculate the feature entropy by
\begin{equation}
    \begin{aligned}
    H_{f}=&logZ-\frac{1}{Z}\sum_{ij}e^{\langle v_{i},v_{j}\rangle}\langle v_{i},v_{j}\rangle\\
    =&logN^2+logE_{v_{i},v_{j}}(e^{\langle v_{i},v_{j}\rangle})-\frac{E_{v_{i},v_{j}}(e^{\langle v_{i},v_{j}\rangle}\langle v_{i},v_{j}\rangle)}{E_{v_{i},v_{j}}(e^{\langle v_{i},v_{j}\rangle})}.
    \end{aligned}
\end{equation}
However, $\langle v_{i},v_{j}\rangle$ is hard to calculate directly in Eq.~(7). In the experiments of graph representation learning, we observe that the absolute value distribution of the values of each dimension is uniform. Therefore, we assume that the absolute value of each element is 1, which is called the distributed hypothesis~\cite{Sahlgren2008The}. Furthermore, to calculate $\langle v_{i},v_{j}\rangle$, we map node embeddings to the $n$-dimensional hyper-sphere with radius $\sqrt{n}$. Then, $\langle v_{i},v_{j}\rangle=n*cos\theta$~\cite{DBLP:conf/emnlp/LiZHWYL20}, where $\theta$ is the angle between any two vectors, and it connects the node embedding dimension $n$ with entropy. Next, the probability distribution of the angle $\theta$ between two random vectors in the $n$-dimensional hyper-sphere is needed to keep $n$ as the only variable. Because of the isotropy, we only need to consider the unit vector. Besides, we only need to fix one of the vectors and consider the random change of the other vector. Therefore, we set the $n$-dimensional random vector $x=(x_{1},x_{2},\cdots,x_{n})$, the fixed vector is $y=(1,0,\cdots,0)$ \cite{WinNT19}. Converting $x$ to hyper-sphere coordinates is defined as:
\begin{equation}
\begin{aligned}
\left\{\begin{array}{llcll}
 x_{1}=r*cos(\varphi_{1})\\ 
 x_{2}=r*sin(\varphi_{1})cos(\varphi_{2})\\ 
\qquad\vdots \\ 
x_{n-1}=r*sin(\varphi_{1})sin(\varphi_{2})\cdots sin(\varphi_{n-2})cos(\varphi_{n-1})\\ 
 x_{n}=r*sin(\varphi_{1})sin(\varphi_{1})\cdots sin(\varphi_{n-2})sin(\varphi_{n-1}),
\end{array}\right.
\end{aligned}
\end{equation}
where $r=\sqrt{x_{1}^{2}+x_{2}^{2}+\cdots+x_{n}^{2}}=1$ is a radial coordinate, $\varphi_{n-1}\in [0,2\pi)$, others $\varphi\in [0,\pi]$. Note that $\|x\|=1$ and $\|y\|=1$, therefore the angle between $x$ and $y$ is
\begin{equation}
    arccos(\frac{x^T y}{\|x\| \|y\|}) = arccos(cos(\varphi_{1}))=\varphi_{1}.
\end{equation}
where $arccos$ is the inverse $cosine$ ($arccosine$) function. Eq.~(9) shows that the angle between $x$ and $y$ is $\varphi_{1}$. In particular, the probability of the angle between $x$ and $y$ that does not exceed $\theta$ is 
\begin{equation}
  \begin{aligned}
       P(\varphi_{1}\leq \theta) 
    = \frac{\Gamma (\frac{n}{2})}{\Gamma (\frac{n-1}{2})\sqrt\pi}\int_{0}^{\theta}sin^{n-2}\varphi_{1}d(\varphi_{1}).
  \end{aligned}
\end{equation}
The probability density of $\theta$ is denoted (following~\cite{henderson1996experiencing}) by
\begin{equation}
    P_{n}(\theta)=\frac{\Gamma (\frac{n}{2})}{\Gamma (\frac{n-1}{2})\sqrt\pi}sin^{n-2}\theta.
\end{equation}
Therefore, plugging the probability density from Eq.~(11) to Eq.~(7), the feature entropy is updated by
\begin{equation}
    H_{f}=logN^2+logE(e^{ncos\theta}))-\frac{E(e^{ncos\theta} ncos\theta)}{E(e^{ncos\theta})},
\end{equation}
where the calculation of the expectation operation is
\begin{gather}
         E(e^{ncos\theta} ncos\theta)=\int_{0}^{\pi}e^{ncos\theta} ncos\theta P_{n}(\theta)d_{\theta},\\ 
    E(e^{ncos\theta})=\int_{0}^{\pi}e^{ncos\theta}P_{n}(\theta)d_{\theta}.
\end{gather}
\paragraph{Structure Entropy.}
Structure entropy measures the complexity of networks, which is original from the Shannon's 1953 question~\cite{DBLP:journals/jacm/Brooks03}. For graph data, we design a novel structure entropy to measure the information of link structures as shown in Fig.~\ref{fig1}. The structure entropy uses normalized node degree as basic unit to further search the ideal dimension, which considers two-hop neighbors. First, given a graph with the adjacency matrix $A$ that contains first-order link structures, the second-order adjacency matrix is defined as $A^{2} = A^{T}A$ to measure the second-order link structures. At the same time, the degree vector $D$ is generated by the adjacency matrix $A$. Second, the normalized degree vector $D_{r}$, which contains first-order and second-order link structures, can be defined as:
\begin{equation}
    D_{r} = D^{T}A_{r}^{2},
\end{equation}
where $A_{r}^{2}$ represents the normalized second-order adjacency matrix defined as:
\begin{gather}
    A_{r}^{2}[i,j] =\frac{A^{2}[i,j]}{\sum_{j}A^{2}[i,j]},
\end{gather}
where $A^{2}[i,j]$ is the value of the $i$-th row and $j$-th column of the second-order adjacency matrix $A^{2}$. Following the paradigm of information entropy, the structure entropy with the normalized degree of nodes as a unit event is defined as:
\begin{align}
    H_{s} &= -\sum_{i}^{N}P_{i}logP_{i}= -\sum_{i}\frac{D_{r}[i]}{\sum_{i}D_{r}[i]}log(\frac{D_{r}[i]}{\sum_{i}D_{r}[i]}),
\end{align}
where $D_{r}[i]$, calculated by Eq.~(14), is the normalized degree of node $i$. Therefore, the graph entropy is defined as:
  \begin{align}
  H_{g}&= H_{f} + \lambda H_{s}\\ \notag
   & = logN^2+log\int_{0}^{\pi}e^{ncos\theta}P_{n}(\theta)d_{\theta}\\ \notag
  &-\frac{\int_{0}^{\pi}e^{ncos\theta} ncos\theta  \notag P_{n}(\theta)d_{\theta}}{\int_{0}^{\pi}e^{ncos\theta}P_{n}(\theta)d_{\theta}} \\ \notag
    &-\lambda \sum_{i}\frac{D_{r}[i]}{\sum_{i}D_{r}[i]}log(\frac{D_{r}[i]}{\sum_{i}D_{r}[i]}). \notag
    \end{align}
\begin{algorithm}[t]
\caption{MinGE}
\LinesNumbered 
\KwIn{Graph $G(V,E,A)$; Hyperparameter $\lambda$;}
\KwOut{The ideal node embedding dimension $n$ for the graph G}
Initialize $A^{2} \xleftarrow{} A^{T}A$;\\
Calculate $A_{r}^{2} \xleftarrow{} Eq.~(16)$;\\
Estimate the feature Entropy by $H_{f} \xleftarrow{} Eq.~(7)$;\\
Calculate $D_{r} \xleftarrow{} Eq.~(15)$;\\
Calculate the structure entropy by $H_{s} \xleftarrow{} Eq.~(17)$; \\
Calculate the graph entropy by $H_{g} \xleftarrow{} Eq.~(18)$; \\
Obtain the ideal node embedding dimension $n \xleftarrow{} H_{g}=0$.\\
\end{algorithm}
\paragraph{Algorithm.} We summarize the whole process of MinGE in Algorithm 1. To be specific, we first perform sampling to approximate the sum operation and map the node embedding to the $n$-dimensional hyper-sphere with radius $\sqrt{n}$ to estimate the feature entropy $H_f$ by Eq.~(7) and (12). Then we calculate the structure entropy $H_s$ by Eq.~(17) based on the normalized degree vector in Eq.~(15). Finally, according to the minimum entropy principle~\cite{zhu1997minimax}, we can get the ideal node embedding dimension $n$ by setting $H_{g}=0$. The calculated dimension $n$ is taken as the ideal embedding dimension that can carry all information of node features and link structures. 
\paragraph{Time complexity analysis.} In theory the time complexity of MinGE is $O(n^{2})$, due to matrix multiplication. However, graphs are usually sparse in practice, and we optimize MinGE with sparse matrix computations, which results in rather efficient computation much less than $O(n^{2})$ in practice (as analyzed in Fig.~\ref{fig3}).


\section{Experiments}
\begin{table}[th]
 \centering
  \small
\begin{spacing}{1}
\begin{tabular}{C{60pt}|C{25pt}C{25pt}C{25pt}C{25pt}}
    \hline
    Dataset& Cora& Citeseer&Pubmed&Airport \\ 
    \hline
   \# Nodes  & 2708 & 3327&19717&3188 \\
    \# Edges& 5429 & 4732&44338&18631\\
    \# Features&1433&3703&4500&4\\
    \# Classes&7&6&3&4\\
    \# Validation Node &500 &500& 500&500 \\
    \# Test Nodes & 1000 & 1000&1000&1000 \\
    \hline
    \end{tabular}
    \end{spacing}
    \caption{Summary of datasets used in our experiments.}
    \label{table1}
\end{table}
\begin{table*}[!h]
 \centering
  \small
\begin{spacing}{1.1}
\begin{tabular}{C{25pt}|C{11pt}C{11pt}C{11pt}C{11pt}C{20pt}C{25pt}C{11pt}C{11pt}C{15pt}|C{15pt}C{11pt}C{11pt}C{11pt}C{25pt}C{11pt}C{11pt}C{11pt}C{15pt}}
    \hline
    Task&\multicolumn{18}{c}{\textbf{Node Classification}}\\
    \hline
    Dataset&\multicolumn{9}{c|}{Cora}&\multicolumn{9}{c}{Citeseer} \\ 
    \hline
    Dim. & 20& 40&60&80&Apt~(98)&120&140&160&180 &20& 40&60&80&Apt~(101)&120&140&160&180\\
    \hline
    MLP&54.5&59.8&57.3&57.3&\textbf{60.6}&59.3&58.3&58.2&60.0&58.8&60.0&59.7&	59.2&\textbf{60.3}&60.0&58.5&59.2&59.3\\
    GCN&82.5&83.2&83.1&83.1&\textbf{83.5}&82.6&83.0&82.9&82.8&64.7&66.2&67.0&67.2&67.4&\textbf{67.6}&67.0&67.4&67.2\\
    GAT&82.9&	83.9&	82.4&	83.1&\textbf{84.3}&	82.8&	83.9&	80.4&	82.8&68.9&	69.5&	68.5&	67.9&\textbf{69.6}&	69.4&	69.4&	68.6&	69.1\\
    GCNII&83.9&	84.3&	84.8&	84.3& \textbf{85.1}&85.0&84.6&	83.9&	84.6&72.4&	72.6&	72.4&	72.4&\textbf{73.5}&72.0&72.9&72.3&71.4\\
    \hline
     Dataset&\multicolumn{9}{c|}{Pubmed}&\multicolumn{9}{c}{Airport} \\ 
    \hline
     Dim. &20&	40&	60&	80&	100&	Apt~(123)&	140&	160&	180&20& 40&60&80&Apt~(100)&120&140&160&180\\
     \hline
     MLP&71.1&72.3&71.5&72.8&72.5&\textbf{74.5}&72.9&73.3&72.5 &50.2&	48.6&	45.7&	47.7&\textbf{54.1}&53.6&	51.2&	52.6&	52.0 \\
     GCN&78.0&78.2&	77.5&	78.8&78.4&\textbf{79.2}&79.0&79.2&	77.9&63.6&	63.9&	64.3&	64.8&\textbf{65.8}&	64.9&	64.0&	64.4&	64.2\\
     GAT&77.4&	77.5&78.9&77.5&77.6&\textbf{79.6}&	79.1&77.8&78.5&65.3&64&65.9&63.9&\textbf{67.9}&67.8&65.1&66.8&67.3\\
     GCNII&78.6&	78.8&	79.5&80.0&79.5&\textbf{80.5}&	80.0&	79.8&	79.7&65.5&	67.9&64.1&	68.8&\textbf{70.4}&	68.9&67.1&	69.2&	68.6\\
    \hline
    Task&\multicolumn{18}{c}{\textbf{Link Prediction}}\\
    \hline
    Dataset&\multicolumn{9}{c|}{Cora}&\multicolumn{9}{c}{Citeseer} \\ 
    \hline
    Dim. & 20& 40&60&80&Apt~(98)&120&140&160&180 &20& 40&60&80&Apt~(101)&120&140&160&180\\
    \hline
    MLP&85.8&88.3&	88.1&90.3	&\textbf{90.7}&	89.8&	90.4&	89.1&90.0&90.1&90.2&90.3&	91.9&	\textbf{92.5}&	91.3&	91.6&	91.8&	92.2\\
    GCN&87.6&90.4&91.3&90.5&\textbf{92.6}&92.1&92.4&87.6&87.7 &90.6&91.9&92.7&	92.9&	\textbf{95.4}&	94.7&	94.5&	94.4&	95.4\\
    GAT&91.1&	93.4&	92.8&	93.5&\textbf{94.4}&	94.1&	94.3&	93.7&	94.2&96.5&97.2&97.1&97.5&	\textbf{97.7}&	97.5&	97.3&	97.5&	97.4\\
    GCNII&94.1&	95.8&	95.7&	96.1&\textbf{96.6}&	96.4&	96.2&	95.9&	96.2&98.9&	99.2&	99.3&	99.5&99.7&	\textbf{99.8}&	99.5&	99.5&	99.6\\
    \hline
    Dataset&\multicolumn{9}{c|}{Pubmed}&\multicolumn{9}{c}{Airport}\\ 
    \hline
    Dim. &20&	40&	60&	80&	100&	Apt~(123)&	140&	160&	180&20& 40&60&80&Apt~(100)&120&140&160&180\\
    \hline
    MLP&85.2&	90.8&	91.0&	92.3&	92.4&\textbf{92.6}&	92.4&	92.3&	91.9&89.6&89.8&89.6&89.9&\textbf{90.2}&89.9&89.4&89.9&89.8\\
    GCN&90.3&	93.1&	93.8&	93.4&	93.7&\textbf{94.1}&	93.7&	94.0&	93.9&89.1&91.8&92.9&92.0&\textbf{93.2}&92.6&92.6&93.1&92.8\\
    GAT&87.0&	89.9&	91.2&	90.9&	90.8&\textbf{92.7}&	90.8&90.9&	92.5&91.8&92.6&92.5&\textbf{92.9}&\textbf{92.9}&92.8&92.7&92.7&92.6\\
    GCNII&95.4&97.6&	98&	98.1&	98.3&\textbf{98.5}&	98.5&	98.4&	98.2&92.5&93.0&93.1&93.5&\textbf{93.8}&	93.7&	93.5&	93.6&92.8\\
    \hline
    \end{tabular}
    \end{spacing}
    \caption{Performance on node classification and link prediction. \textit{Apt} denotes the appropriate dimensions calculated by MinGE.}
    \label{table2}
\end{table*}
In this section, we conduct extensive experiments on benchmark datasets to demonstrate the effectiveness and generalizability of the proposed MinGE for GNN methods.
\subsection{Experimental Settings}
\paragraph{Datasets.} In our experiments, we choose the most authoritative benchmark datasets, Cora, Citeseer, Pubmed proposed by~\cite{DBLP:journals/aim/SenNBGGE08} and Airport~\cite{DBLP:conf/nips/ChamiYRL19}. More details of the datasets are shown in Table~\ref{table1}.
We conduct node classification and link prediction on above benchmark datasets, and evaluate the performance of our MinGE.

\paragraph{GNNs.}
We validate the effectiveness and generalizability of MinGE on most popular models, including Multi-Layer Perception~(MLP), Graph Convolution Network~(GCN)~\cite{DBLP:conf/iclr/KipfW17}, Graph Attention network~(GAT)~\cite{DBLP:conf/iclr/VelickovicCCRLB18} and Graph Convolution Network via Initial residual and Identity mapping~(GCNII)~\cite{DBLP:conf/icml/ChenWHDL20}.

\paragraph{Protocols.} For node classification task, following the experiment setting of GAT~\cite{DBLP:conf/iclr/VelickovicCCRLB18}, only 20 samples per class are allowed for training. We further choose 80, 140 per class for Cora, Citeseer and Airport, and 2000, 3000 per class for Pubmed to crop the data set at different scales and postpone the index to divide the validation set with 500 and a test set with 1000. For link prediction task, we randomly split the edges with a ratio of 85\%, 5\%, and 10\% for training, validation and test sets. For both tasks, comparative experiments are constructed by setting dimension interval to 20 for GNNs. Early stopping strategy is used on validation set with a patience of 100 epoches for all experiments. All results are the average of 10 times. Furthermore, experiments are conducted to validate the efficiency of MinGE.
\begin{figure*}[!t]
 \subfigtopskip=0pt
 \subfigbottomskip=1pt

 \subfigure[MLP on Cora.]{
 \includegraphics[width=4.3cm,height=3.2cm]{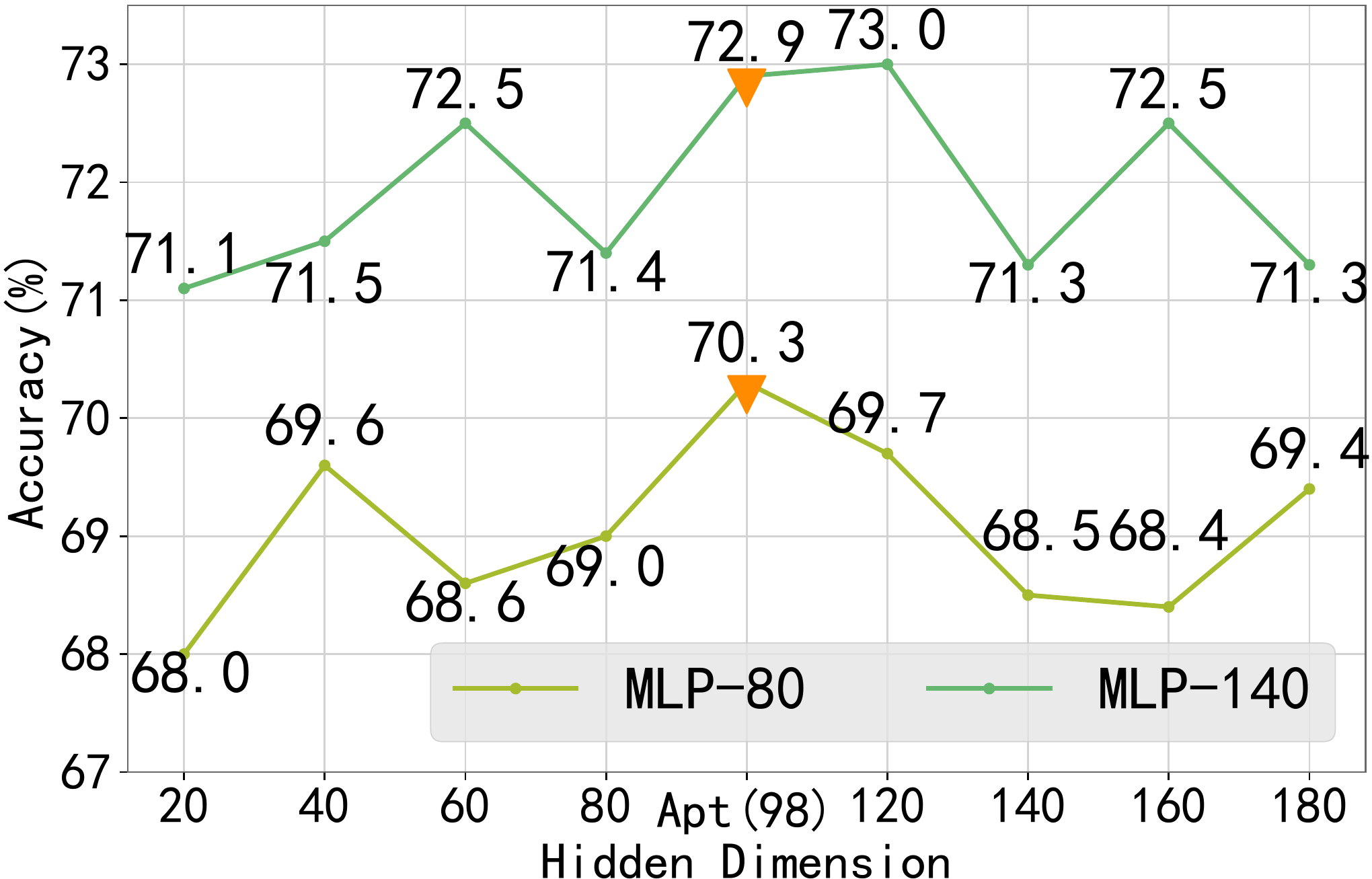}}
 \subfigure[GCN on Cora.]{
 \includegraphics[width=4.3cm, height=3.2cm]{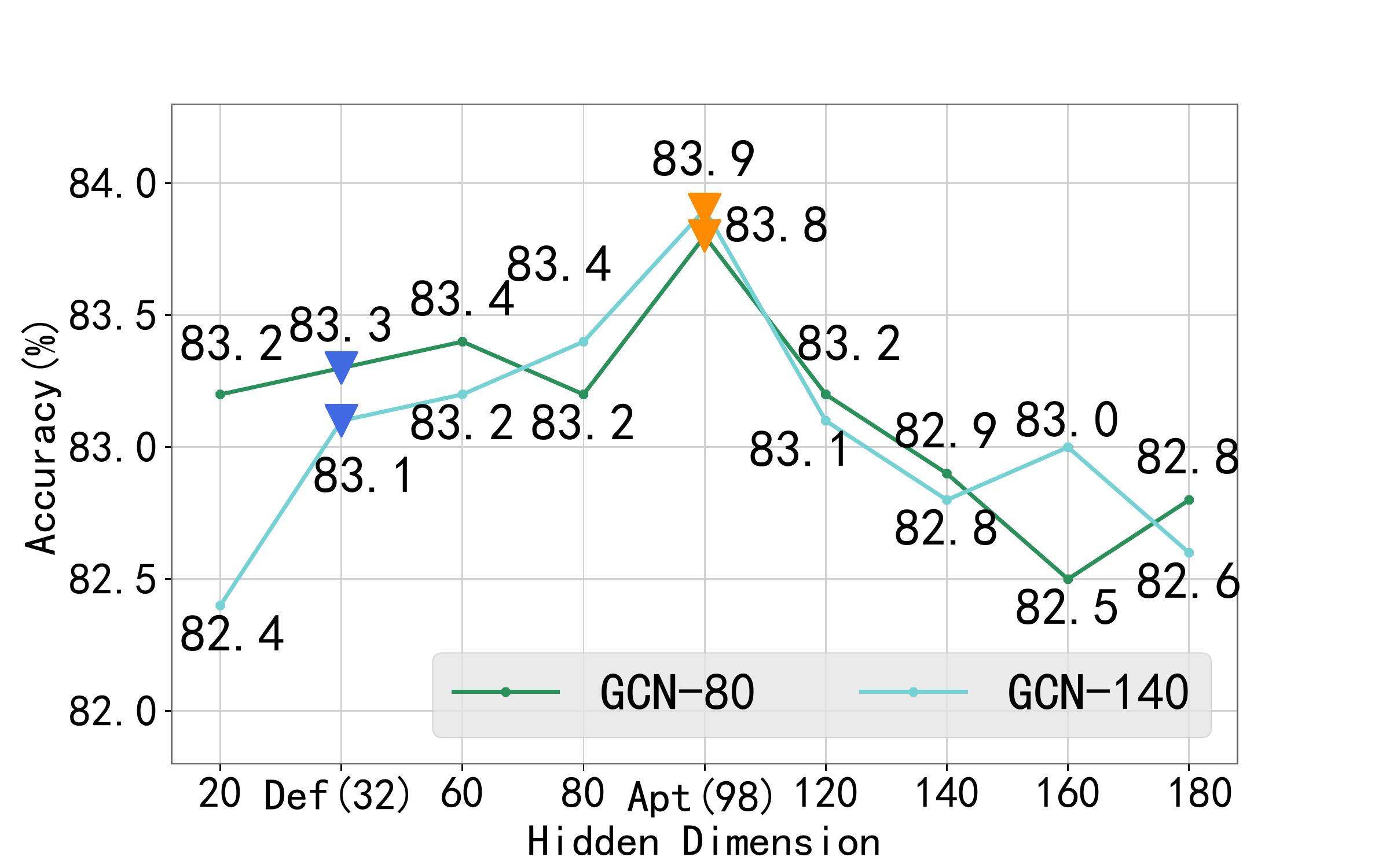}}
 \subfigure[GAT on Cora.]{
 \includegraphics[width=4.3cm, height=3.3cm]{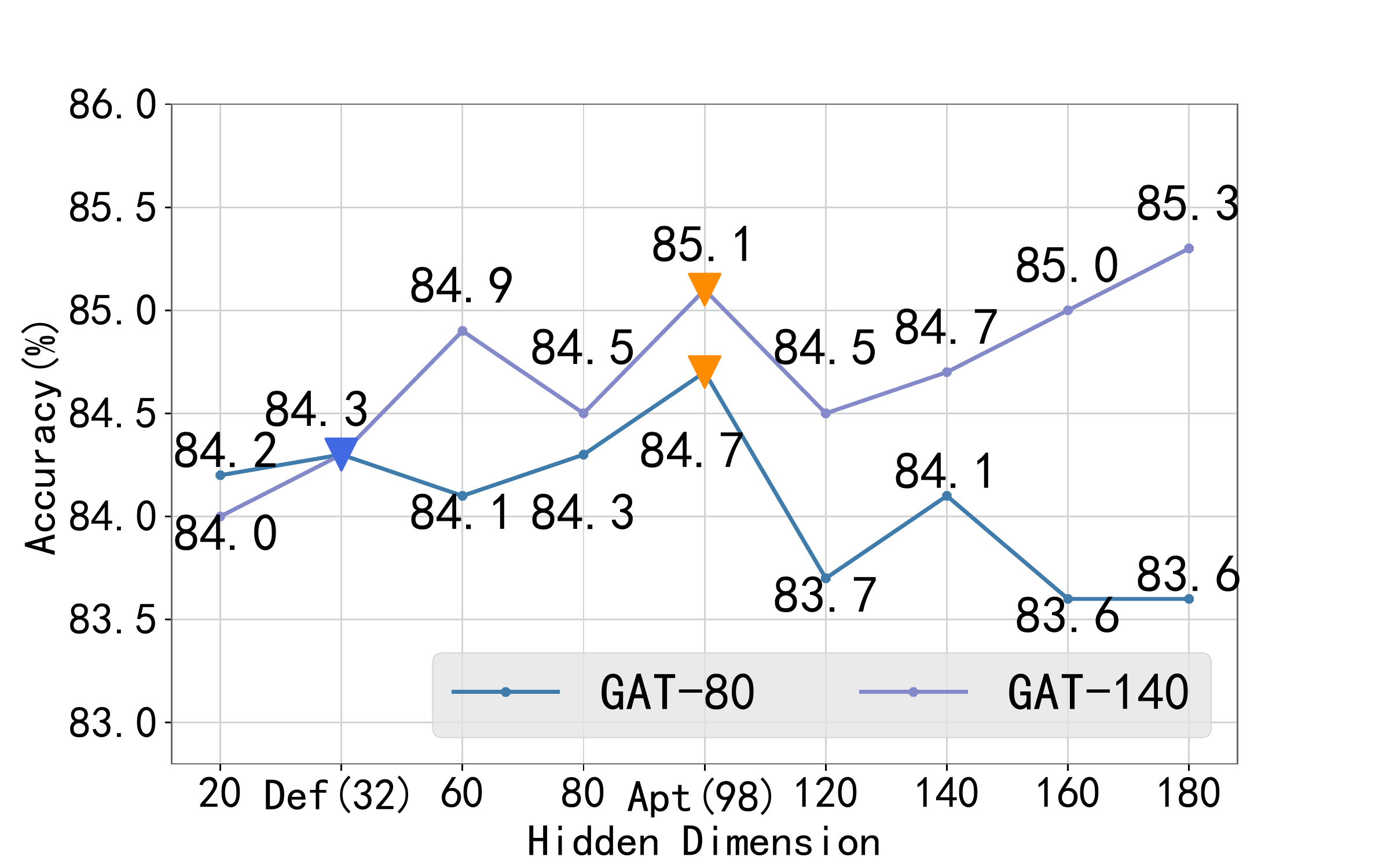}}
 \subfigure[GCNII on Cora.]{
 \includegraphics[width=4.3cm, height=3.2cm]{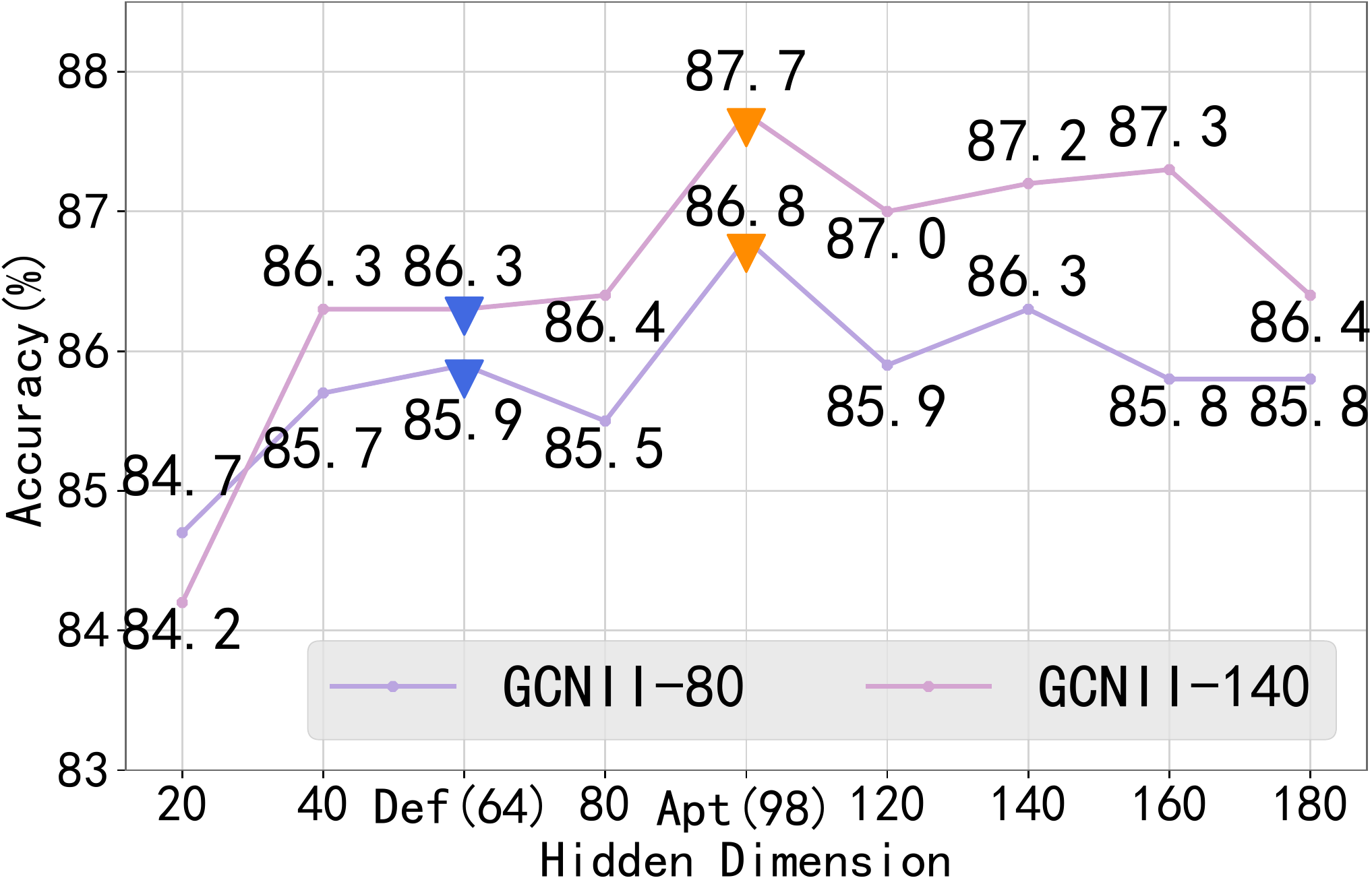}}
 
 \subfigure[MLP on Citeseer.]{
 \includegraphics[width=4.3cm, height=3.2cm]{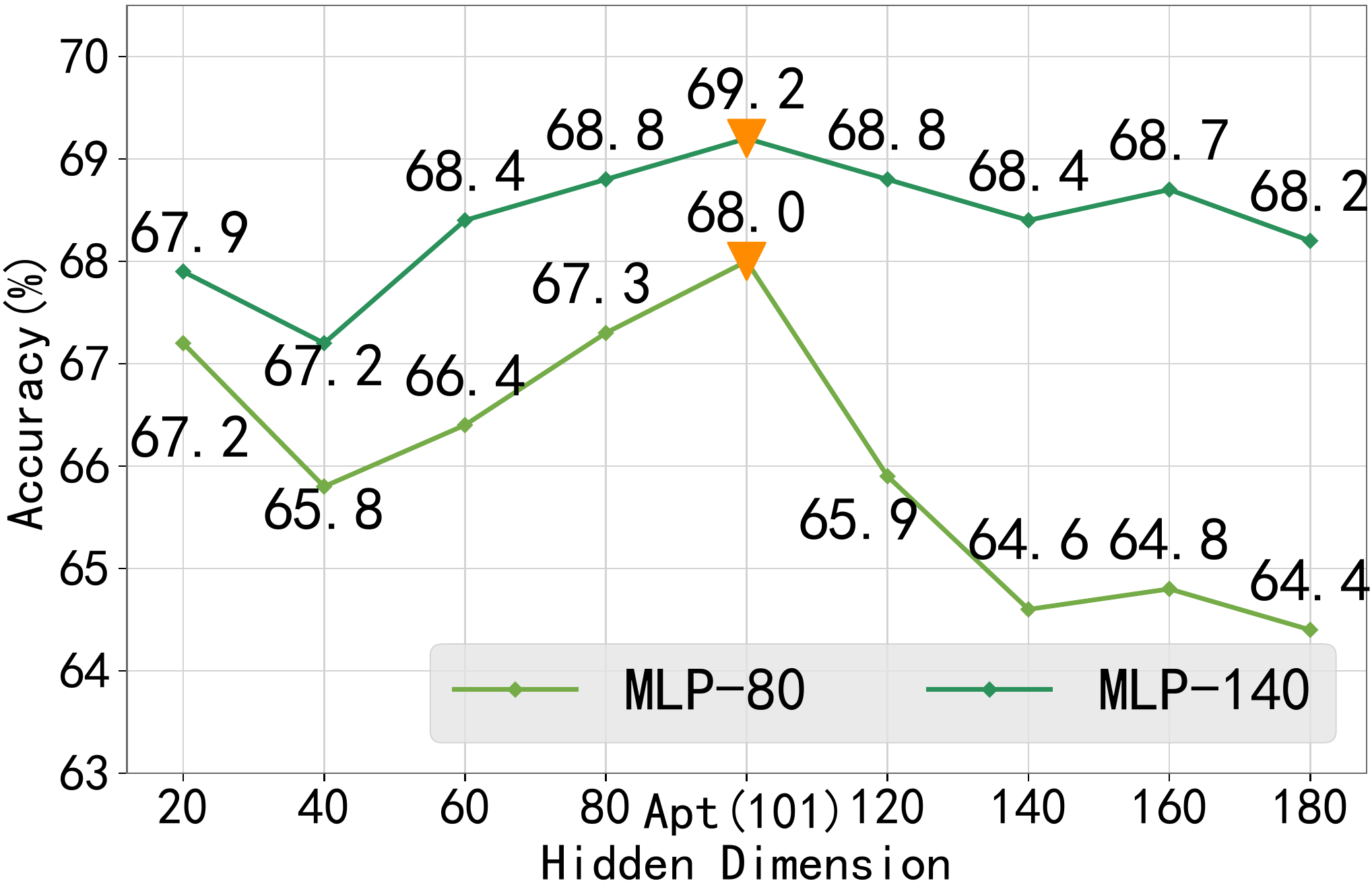}}
 \subfigure[GCN on Citeseer.]{
 \includegraphics[width=4.3cm, height=3.2cm]{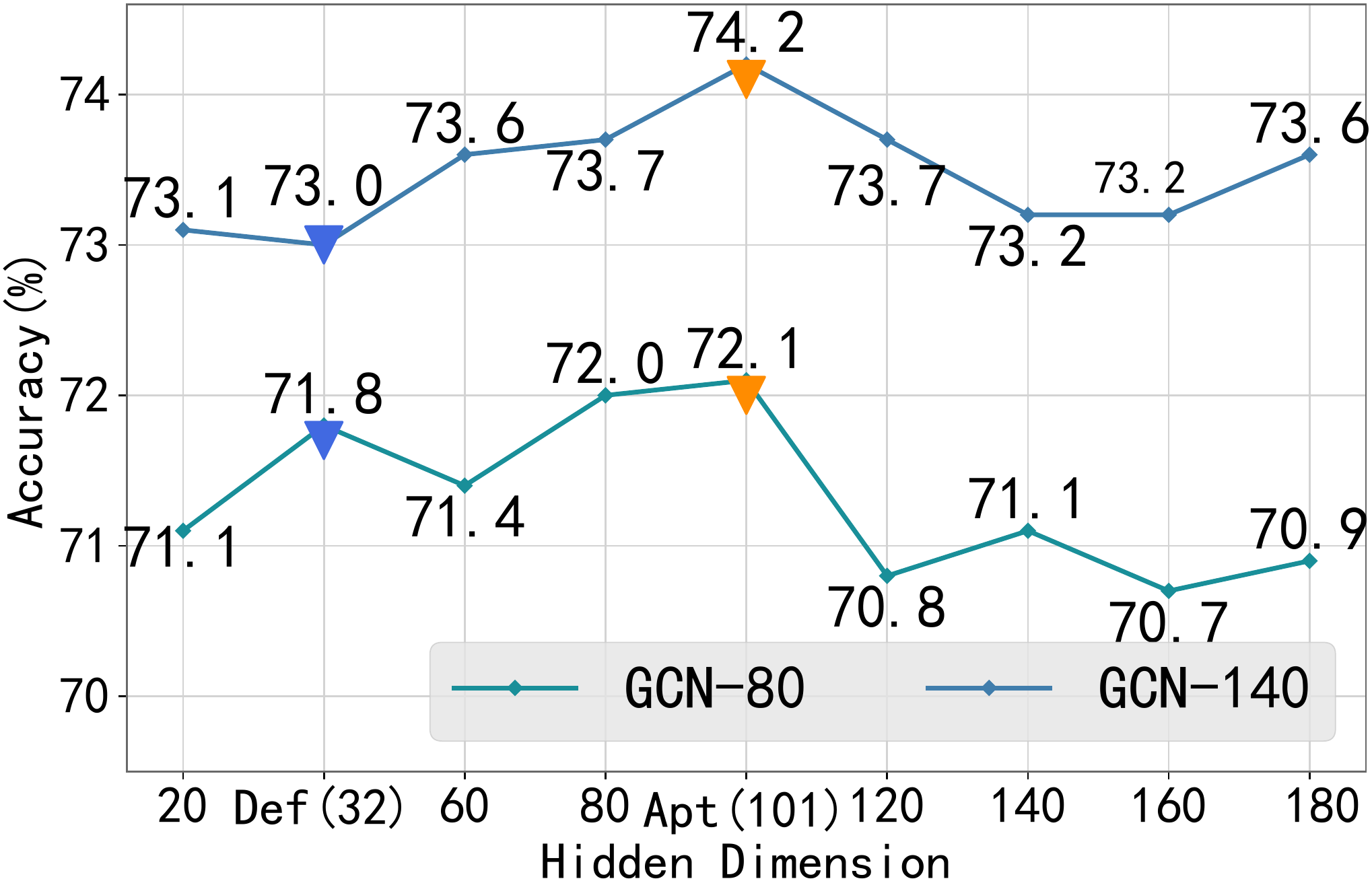}}
 \subfigure[GAT on Citeseer.]{
 \includegraphics[width=4.3cm, height=3.2cm]{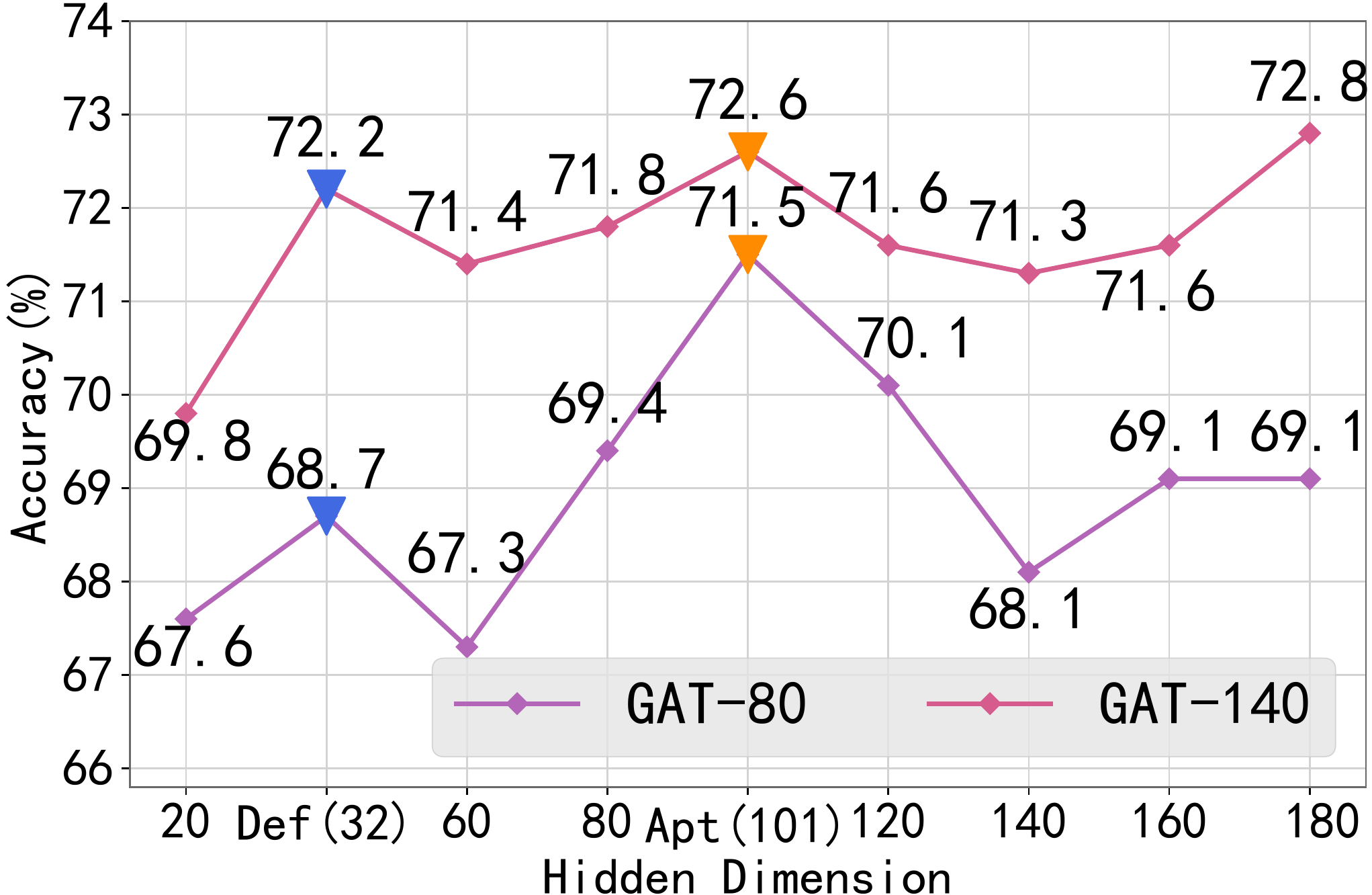}}
 \subfigure[GCNII on Citeseer.]{
 \includegraphics[width=4.3cm, height=3.2cm]{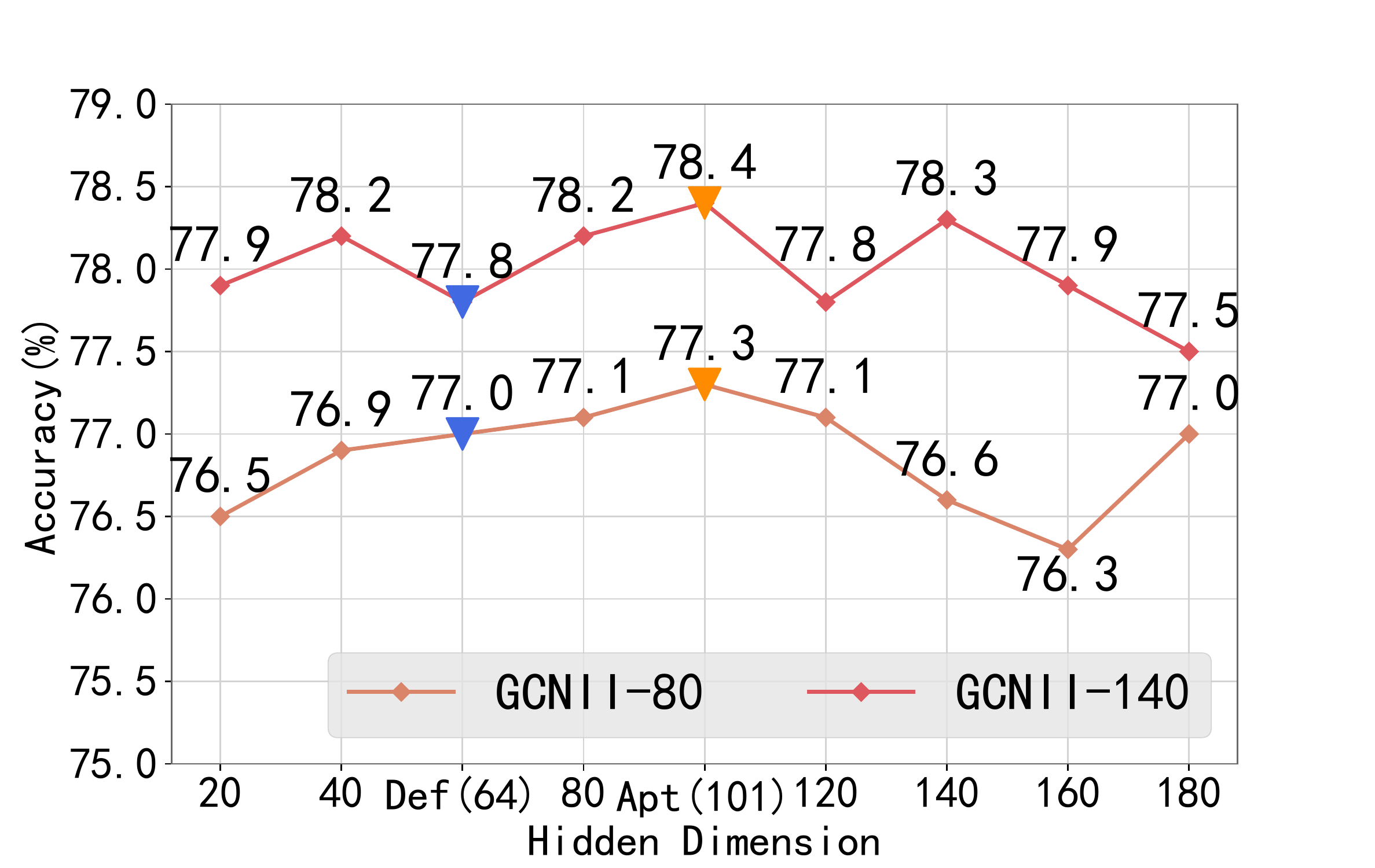}}
 
\subfigure[MLP on Pubmed.]{
\includegraphics[width=4.25cm, height=3.2cm]{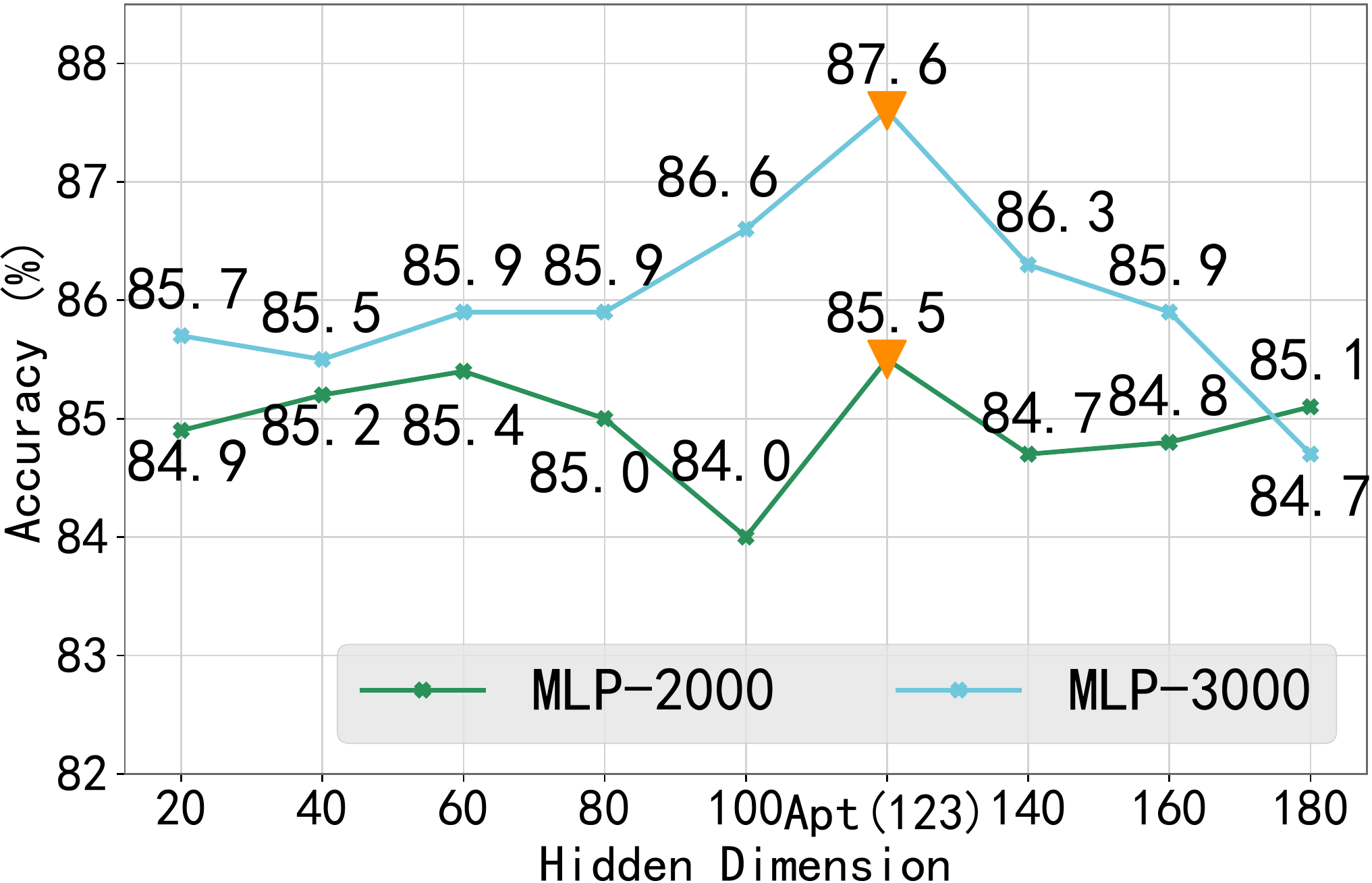}}
\subfigure[GCN on Pubmed.]{
\includegraphics[width=4.25cm, height=3.2cm]{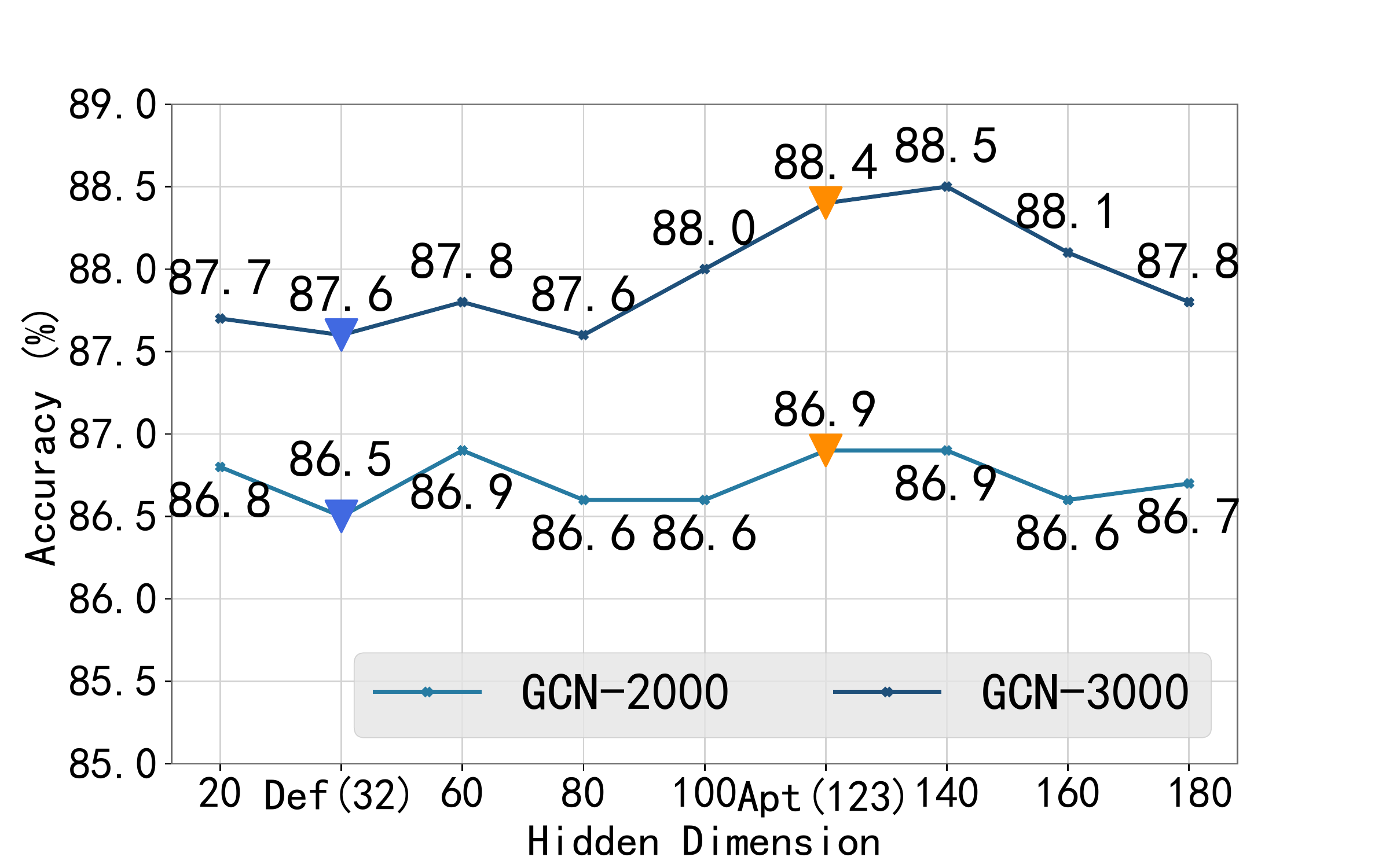} }
\subfigure[GAT on Pubmed.]{
\includegraphics[width=4.25cm, height=3.2cm]{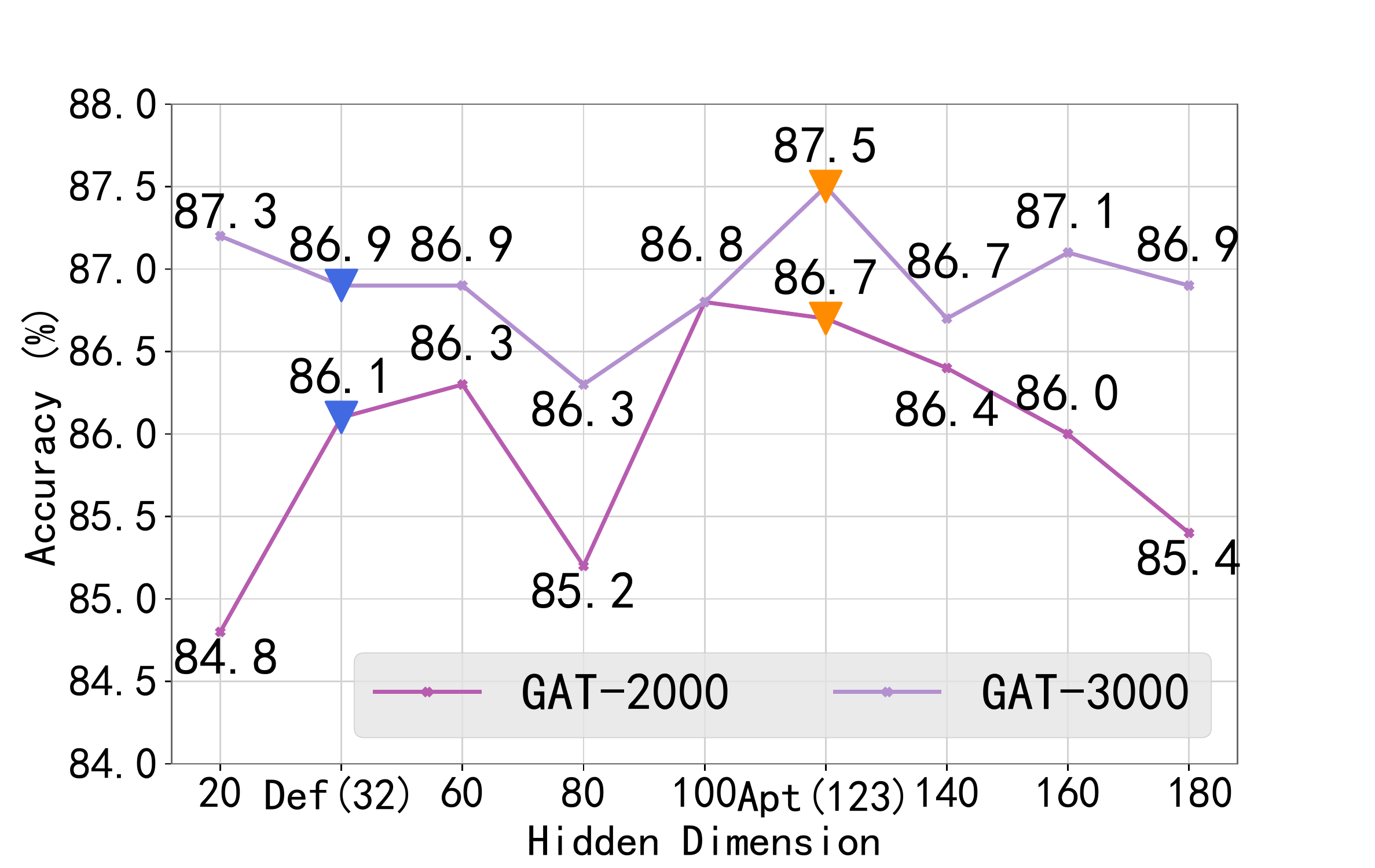}}
 \subfigure[GCNII on Pubmed.]{
 \includegraphics[width=4.3cm, height=3.2cm]{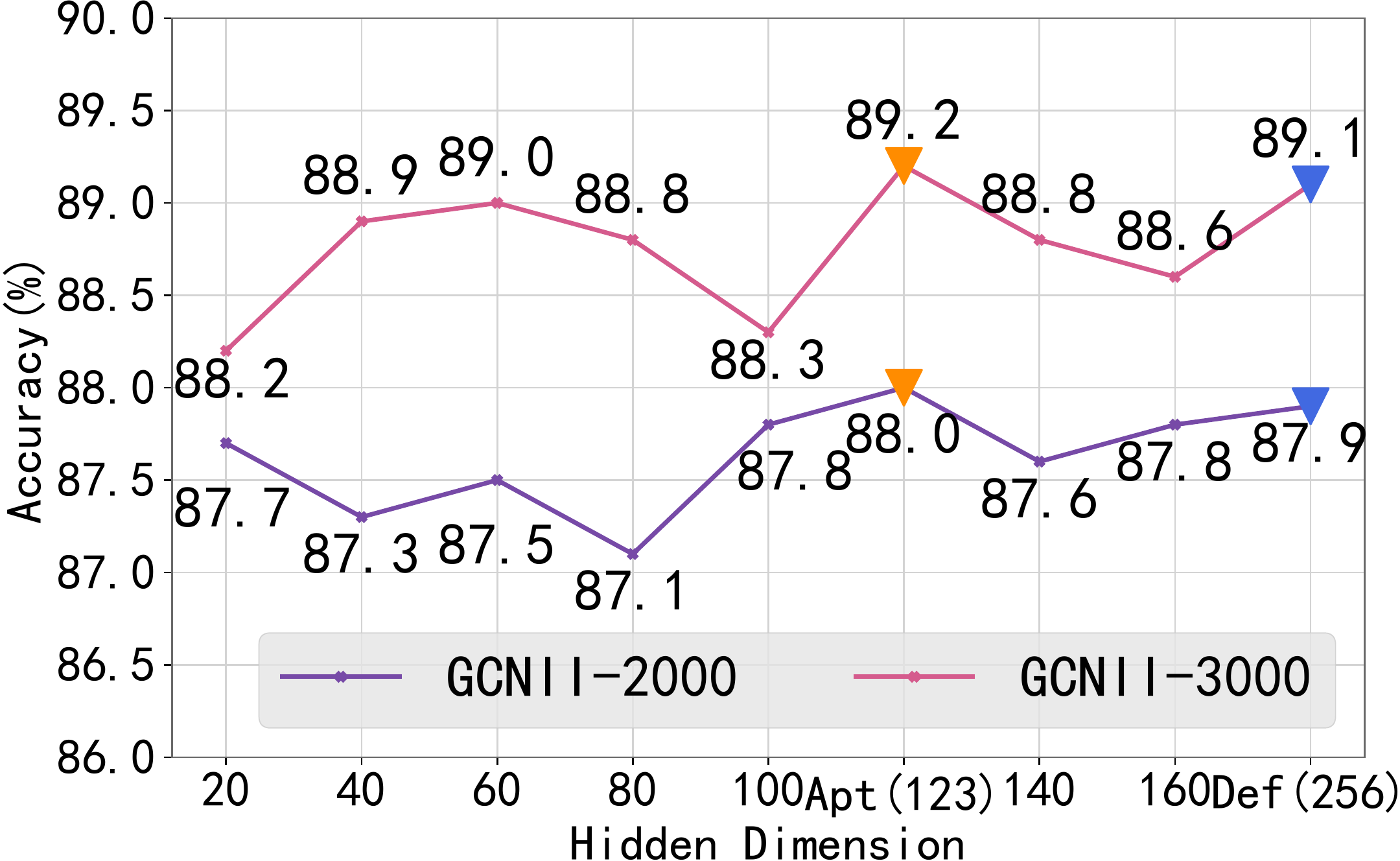} } 
 
\subfigure[MLP on Airport.]{
\includegraphics[width=4.25cm, height=3.2cm]{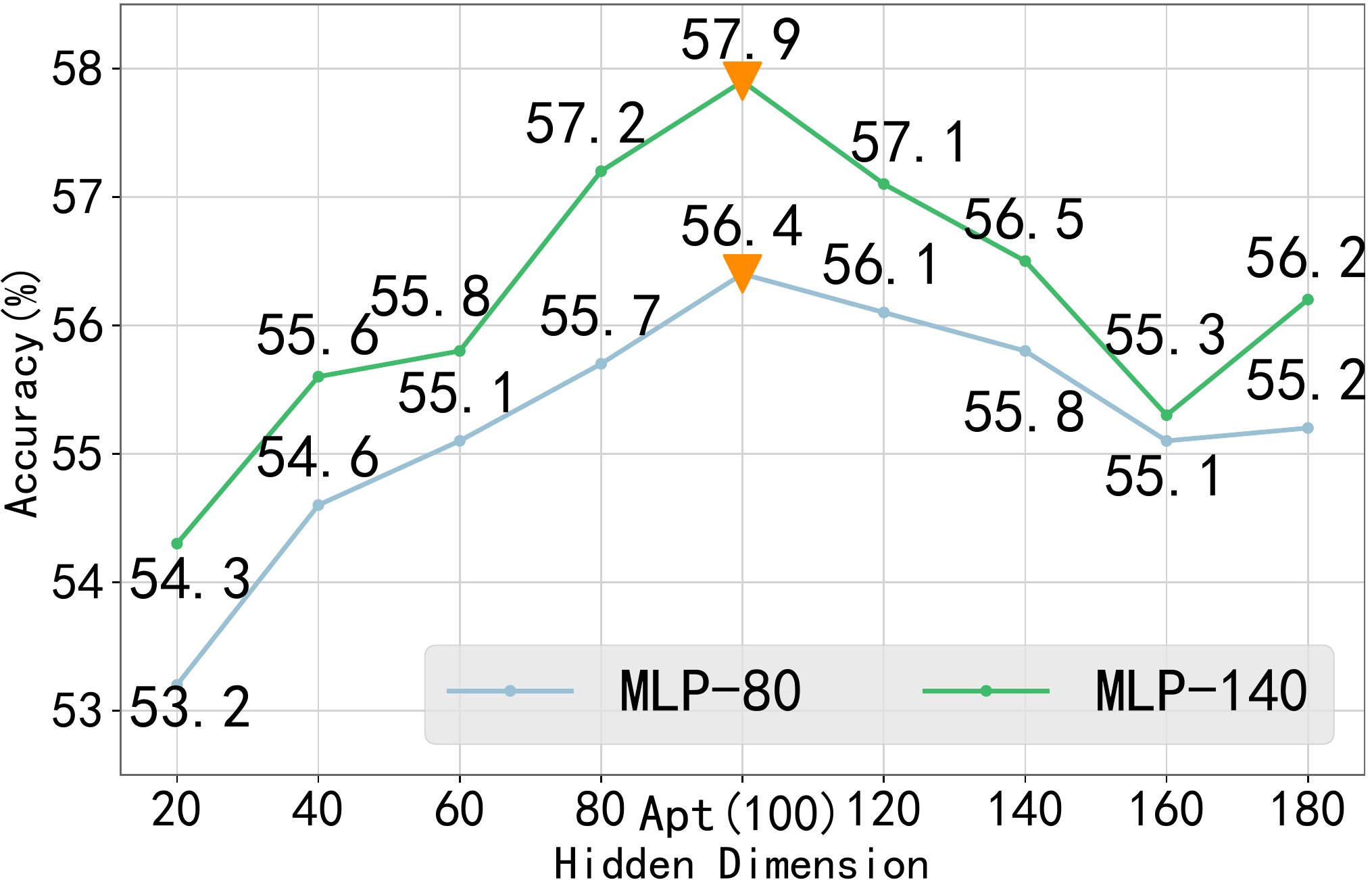}}
\subfigure[GCN on Airport.]{
\includegraphics[width=4.25cm, height=3.2cm]{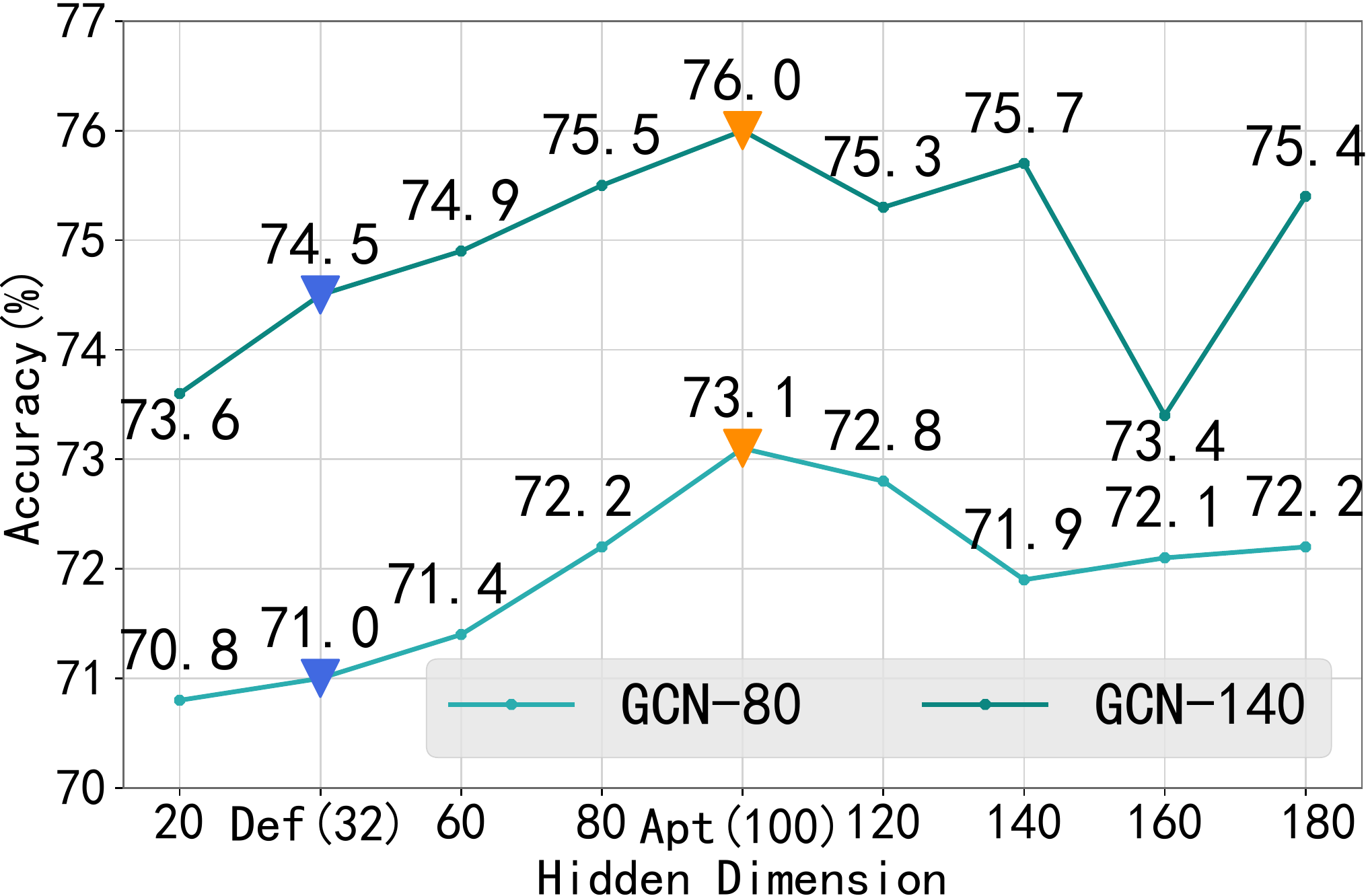} }
\subfigure[GAT on Airport.]{
\includegraphics[width=4.25cm, height=3.2cm]{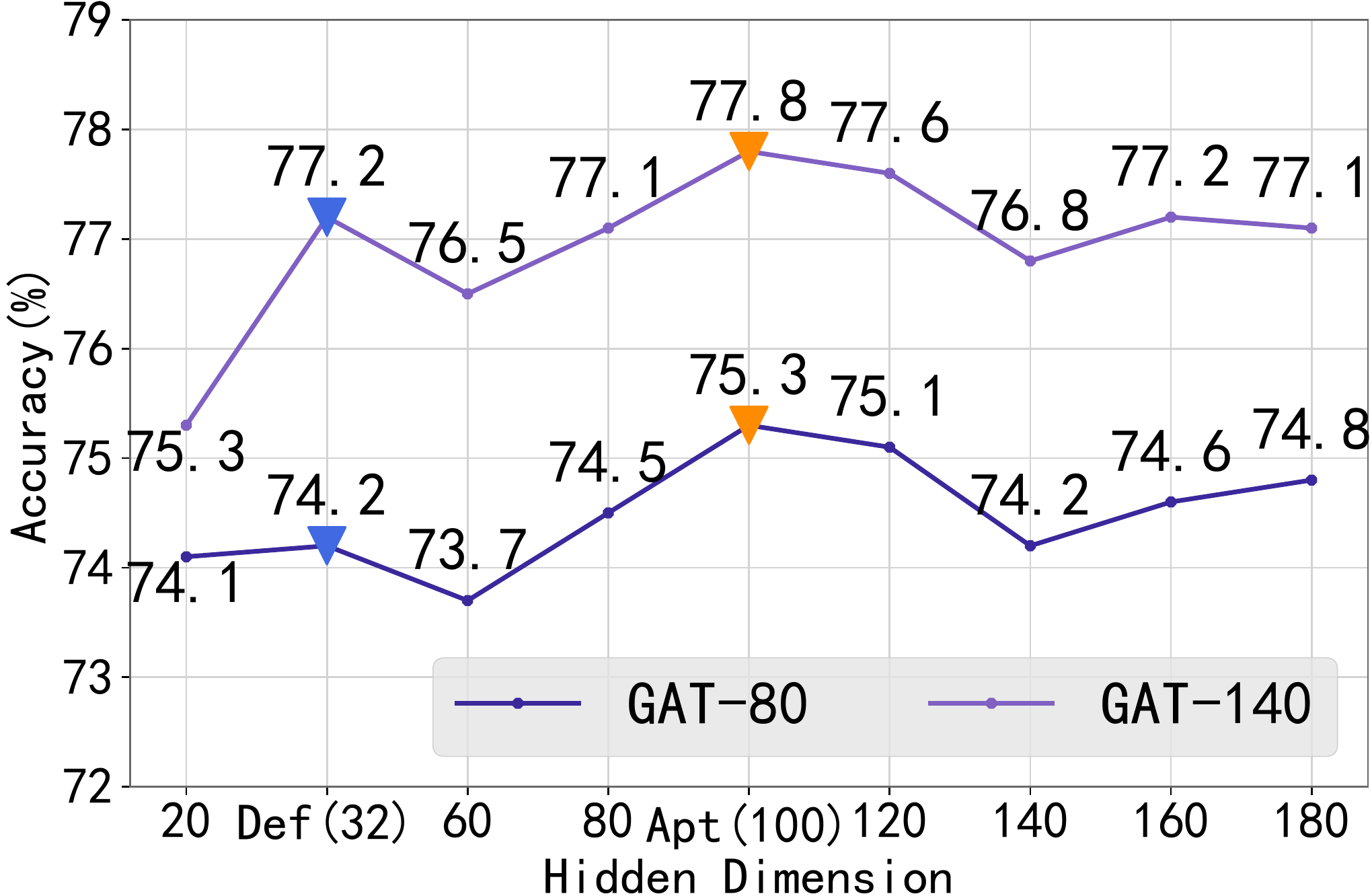}}
 \subfigure[GCNII on Airport.]{
 \includegraphics[width=4.3cm, height=3.2cm]{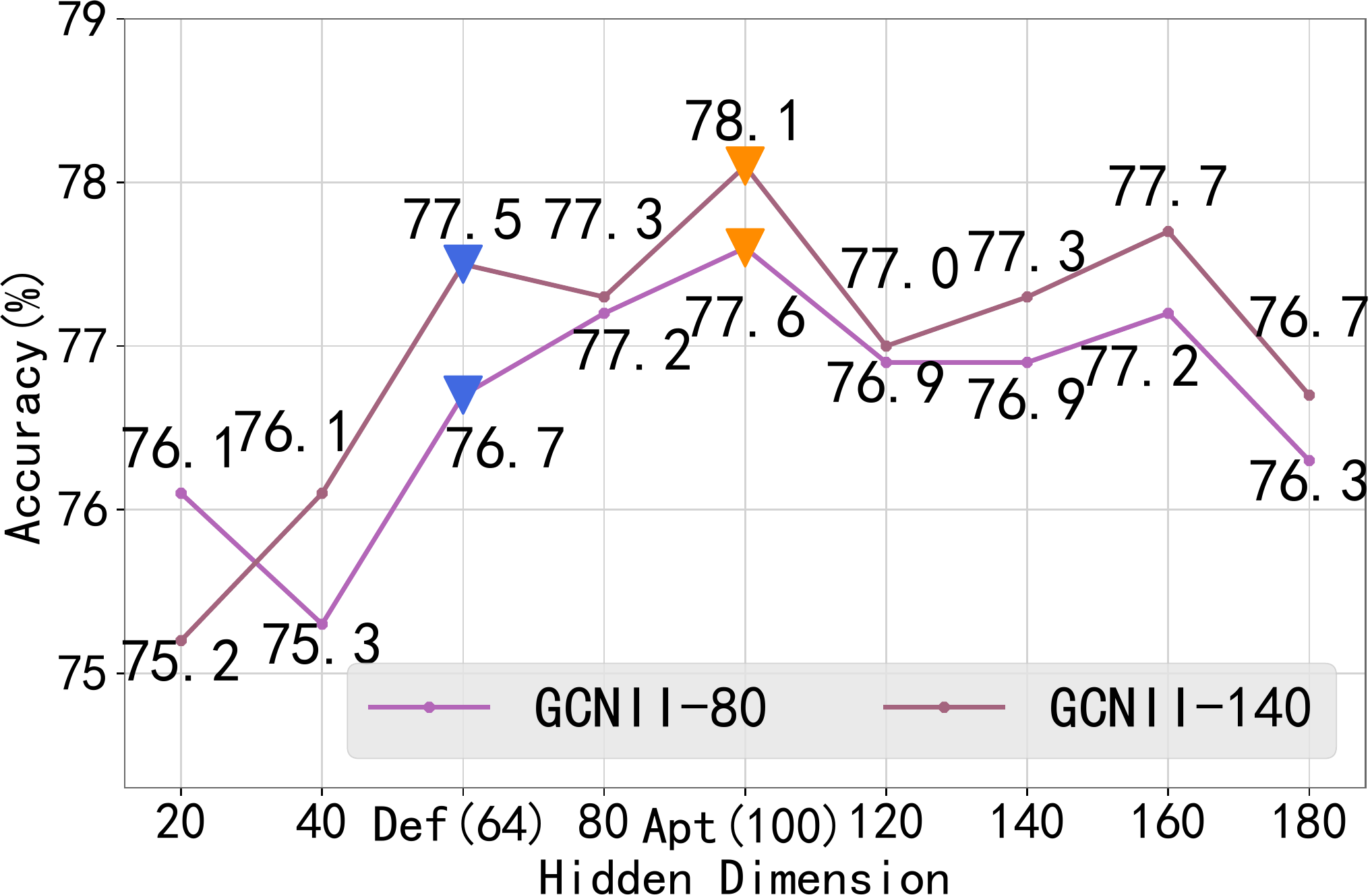} } 

 \caption{Node classification accuracy with different dimensions. Numbers like 80 and 140 in the legends denote the numbers of samples separately selected per class in the training set. The yellow inverted triangles denote the performance of GNNs with the appropriate dimensions selected by MinGE and the blue ones denote that with default dimensions mentioned in the original papers (MLP has no default dimension).}
 \label{fig2}
\end{figure*}
\paragraph{Hyperparameters.} MinGE has only one hyperparameter $\lambda$ to control the ratio of the structure entropy. We set $\lambda=1$ in experiments and analysis it in Section~4.2. The values of node embedding dimension calculated by the MinGE algorithm are 98, 101, 123, 100 for Cora, Citeseer, Pubmed and Airport separately. Where these dimensions are the size of the last layer of GNNs representation model. In practice, we set hidden as a hyperparemeter to control the dimension size. For other hyperparameters, we follow the settings in the original text for most models, which are considered to be optimal. However, because of the differences between the datasets, we also make corresponding adjustments. 
For the MLP model, the optimizer is Adam with learning rate is 0.01 and weight decay is 0.0005, epoch is 250, drop out is 0.5.
For GAT and GCN models, we fully keep the default settings.
For GCNII model, the hyperparameters partly follow the default settings and early stopping strategy. The differences are that the learning rate is 0.008 for Cora, the lamda is 0.3 for Citeseer, the dropout is 0.5 and lamda is 0.4 for pubmed, and layer is 16 for all datasets.
\subsection{Results and Analysis}
\paragraph{Performance on node classification and link prediction.} For node classification and link prediction tasks, the results of GNNs with different dimensions on benchmark datasets are shown in Table~\ref{table2}. Where \textit{Apt} denotes the appropriate dimension calculated by the MinGE in all Tables. From Table~\ref{table2}, by comparing with other dimensions, we can see that GNNs with the appropriate dimension selected from MinGE achieve the best or near the best performance. Besides, from the perspective of overall trend, we can see that it always locates in the peak area of the result change curve. Moreover, we also compare with the experimental results of GCN, GAT, GCNII models in the original texts. We find that GNNs with MinGE can improve 2\%, 1.3\%, 1.6\% on Cora and 0.6\%, 0.6\%, 0.2\% on Pubmed respectively. The above experimental results verify the effectiveness and generalizability of MinGE in guiding the NEDS for GNNs on different tasks and GNN models.
The main reason is that around an appropriate dimension for a GNN model, the lower node embedding dimension is hard to cover all information due to the limitation of expression capacity, and the higher one tends to overfit the training data due to the excessive model complexity. In order to search an appropriate dimension for each model of each task,
MinGE leverages the both feature entropy and structure entropy to measure the node features and link structures in the graph for NEDS, which leads to better performance.

Furthermore, in order to eliminate the influence of uneven data distribution and further verify the genralizability of MinGE, we conduct experiments on benchmark datasets cropped at different scales.
The performance of GNNs with 80, 140 samples per class for Cora, Citeseer and Airport separately and 2000, 3000 samples per class for Pubmed are shown in Fig.~\ref{fig2}. Where \textit{Apt} denotes the ideal dimension selected by MinGE, and \textit{Def} denotes the default dimension mentioned in original paper.
According to the results, it is obvious that the performance of GNNs with the appropriate dimension selected by MinGE always locate in the peak center of the accuracy curves, and the accuracy of each model has been greatly improved as the training set increases. Furthermore, compared with GNNs with default dimension, GNNs with the ideal dimension all achieve better performance. The experimental results are in line with the conclusions verified in Table~\ref{table2}, which indicates generalizability of MinGE.

\begin{table}[t]
 \centering
  \small
\begin{spacing}{1.1}
\begin{tabular}{C{26pt}|C{14pt}C{14pt}C{14pt}C{14pt}C{14pt}C{14pt}C{14pt}C{14pt}}
    \hline
    Layers&2&4&8&16&2&4&8&16\\
    \hline
    Dim.&Def&Def&Def&Def&Apt&Apt&Apt&Apt\\
    \hline
    Cora&80.2&82.3&82.8&83.5&82.5&83.0&83.9&85.1\\
    Citeseer&66.1&67.9&70.6&72.0&72.1&72.7&73.2&73.5\\
    Pubmed&77.7&78.2&78.8&80.3&79.7&79.9&80.0&80.5\\
    Airport&64.1&64.3&65.0&66.7&66.5&67.8&69.2&70.4\\
    \hline
    \end{tabular}
    \end{spacing}
    \caption{Node classification accuracy with varying numbers of GNN (GCNII) layers.}
    \label{table3}
\end{table}
\paragraph{Memory Efficiency.}
GCNII is the current state-of-the-art model that can achieve better performance with deeper neural networks, as shown in Table~\ref{table3}.
However, deeper network spends more memory and time during model training.
Compared with the node dimension by default in original paper,
MinGE can find an appropriate dimension for GCNII with different numbers of layers.
In such a case,
GCNII with MinGE can use fewer layers but achieve similar or better performance to a deeper GCNII with the node dimension by default setting.
For example, the performance of GCNII with 16 layers and the node dimension by default is worse than GCNII with two layers and an appropriate dimension selected by MinGE on Citeseer.
In summary, by using MinGE, GCNII can choose fewer layers to achieve good performance while memory limitation becomes crucial. 

\begin{figure}[b]
  \subfigtopskip=0pt
  \subfigbottomskip=0pt
    \centering
    \includegraphics[width=8.7cm,height=5.1cm]{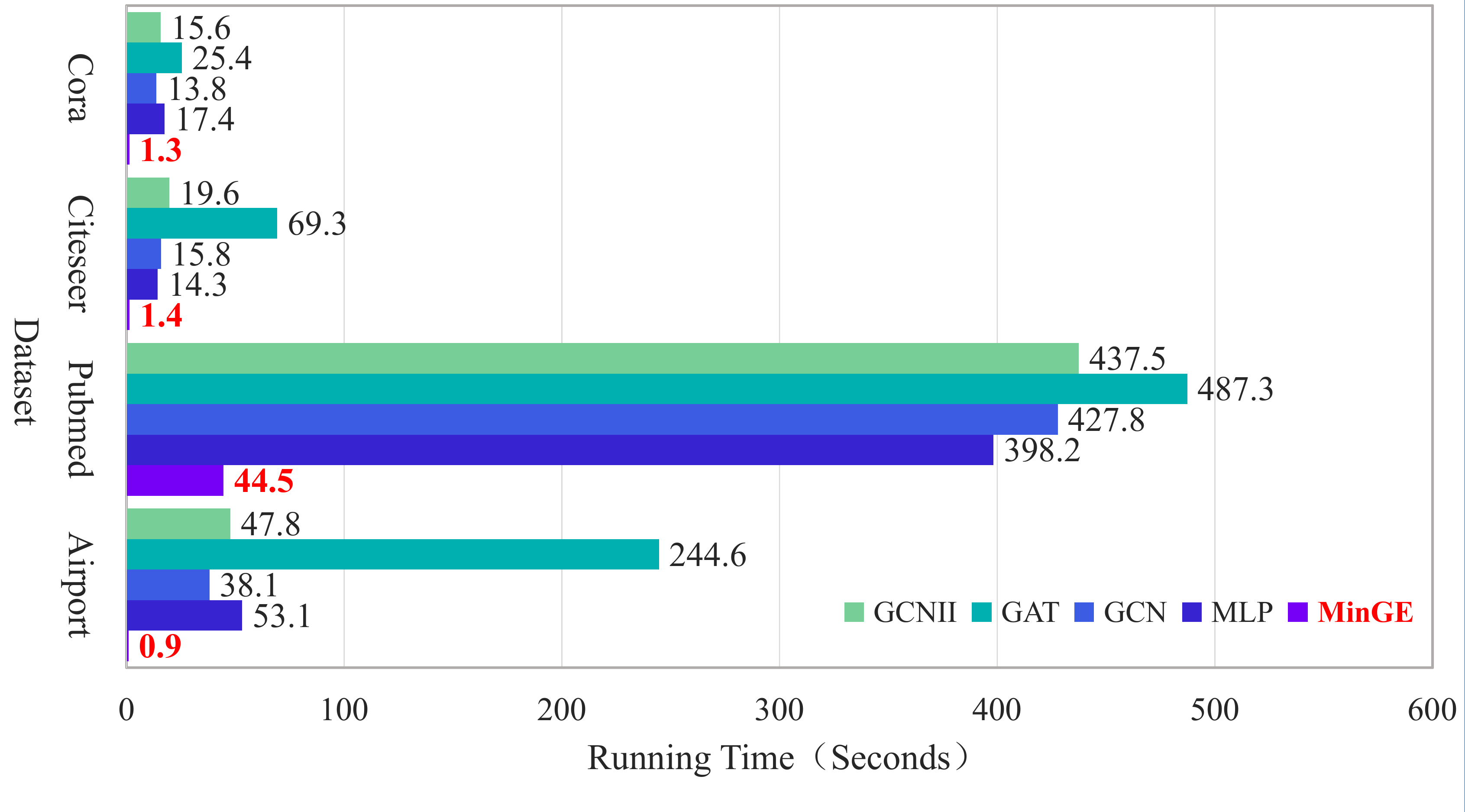}
    \caption{Running time of MinGE compared with GNNs.}
    \label{fig3}
\end{figure}
\paragraph{Time Efficiency.} Considering fairness, we conduct experiments to compare the time that MinGE runs once with that GNNs run once. Fig.~\ref{fig3} shows the running time of MinGE and GNN methods. From these results, we can observe that MinGE costs much less runtime compared with training GNNs. It turns out our approach adds almost no additional computational overhead and is more efficient for an appropriate embedding dimension selection through MinGE. Note that, our approach only requires the runtime of MinGE with a single execution of different GNNs here, while finding a good dimension by grid search needs to run the GNNs multiple times, so as to cost more computational burden and time.

\paragraph{Hyperparameter Analysis.}
In our MinGE algorithm, there is only one hyperparameter $\lambda$, which controls the weight of structure entropy. The structure entropy provides additional help to the feature entropy. In order to determine the weight of the strcture entorpy, we use GCN to analyze $\lambda$ on benchmark datasets. All dimensions calculated by MinGE are rounded up. From Table~\ref{table4}, it is not hard to see that the structure entropy is often equally important as the feature entropy for NEDS in general. Therefore, we can simply set $\lambda$ to 1 by default given an arbitrary graph. We also observe that the graph entropy with $\lambda=0.1$, which can be considered as feature entropy alone, can give pretty good results. Adding structure entropy with different weights can further improve the performance but with slight variance compared with experimental results as shown in Table~\ref{table2}.

 \begin{table}[!t]
 \centering
  \small
\begin{spacing}{1}
\begin{tabular}{C{25pt}|C{14pt}C{14pt}C{14pt}C{14pt}|C{14pt}C{14pt}C{14pt}C{14pt}}
    \hline
    Dataset&\multicolumn{4}{c|}{Cora}&\multicolumn{4}{c}{Citeseer} \\ 
    \hline
    $\lambda$ &0.1 &0.5&1&2&0.1 &0.5&1&2 \\
    Dim& 69 & 82&98&131&71&84&101&134 \\
    GCN&82.6&83.2&\textbf{83.5}&83.2&67.2&67.4&67.4&\textbf{67.6}\\
    \hline
    Dataset&\multicolumn{4}{c}{Pubmed}&\multicolumn{4}{c}{Airport} \\ 
    \hline
    $\lambda$ &0.1 &0.5&1&2&0.1 &0.5&1&2 \\
    Dim&86&102&123&164&71&84&100&133\\
    GCN&78.8&78.4&\textbf{79.2}&79.1&64.5&64.7&\textbf{65.8}&64.3\\
    \hline
    \end{tabular}
    \end{spacing}
    \caption{Node classification accuracy with varying hyper-parameters.}
    \label{table4}
\end{table}
\vspace{-0.2cm}
\section{Conclusion}
In this paper, we revisit Node Embedding Dimension Selection~(NEDS) for graph data from the perspective of minimum entropy principle. We proposed MinGE, a novel algorithm that combines well-designed feature entropy and structure entropy, to guide the NEDS for GNNs and addressed the challenge of how to directly select the appropriate node embedding dimension for graph data. We focused on MinGE that can maximize node information and structure information for appropriate NEDS by using minimum entropy principle. Moreover, to the best of our knowledge, MinGE is the first study that applies minimum entropy theory to NEDS of graph data. In practice, with this theory, we discovered the effectiveness and generalizability of our MinGE algorithm for popular GNNs. All of our discoveries were concretely validated on benchmark datasets. 


\bibliographystyle{named}
\bibliography{ijcai21.bib}
\appendix

\end{document}